\documentclass[11pt]{article}

\usepackage[final]{acl}
\usepackage{times}
\usepackage{latexsym}
\usepackage{microtype} %
\usepackage{inconsolata} %
\usepackage[usenames,dvipsnames,table]{xcolor}
\usepackage{xspace}
\usepackage{amsmath,amssymb}
\usepackage{graphicx}
\usepackage{booktabs}
\usepackage{listings}
\usepackage{cleveref}
\usepackage{soul}
\usepackage{ifthen}
\usepackage{outlines} %
\usepackage{subcaption}
\usepackage{threeparttable}

\usepackage{longtable, booktabs, siunitx, caption}

\usepackage{url}

\usepackage{siunitx}
\usepackage[normalem]{ulem}

\usepackage{longtable}

\newcommand{\newterm}[1]{\textbf{#1}\xspace}

\definecolor{maroon}{RGB}{178,0,0}
\definecolor{turquoise}{RGB}{2,194,208}
\definecolor{yellow}{RGB}{255, 251, 0}
\definecolor{lime}{RGB}{166, 255, 0}
\definecolor{pink}{RGB}{255, 124, 124}

\newcommand{\myMathMacro}[2]{
    \newcommand{#1}{\textcolor{turquoise}{\ensuremath{#2}}\xspace}
}
\myMathMacro{\mi}{\mathrm{MI}}

\newcommand{\myTextMacro}[2]{
    \newcommand{#1}{\textcolor{black}{#2}\xspace}
}
\myTextMacro{\dependentVar}{expressivity}

\myTextMacro{\vft}{simple fine-tuning}
\myTextMacro{\Vft}{Simple fine-tuning}
\myTextMacro{\fft}{full fine-tuning}
\myTextMacro{\Fft}{Full fine-tuning}
\myTextMacro{\asrft}{ASR fine-tuning}
\myTextMacro{\textft}{text fine-tuning}
\myTextMacro{\Textft}{Text fine-tuning}
\myTextMacro{\tkrep}{tokenizer replacement}
\myTextMacro{\Tkrep}{Tokenizer replacement}
\myTextMacro{\mtl}{multitask fine-tuning}
\myTextMacro{\Mtl}{Multitask fine-tuning}
\myTextMacro{\mtlonly}{multitask-only}
\myTextMacro{\Mtlonly}{Multitask-only}
\myTextMacro{\fftwtk}{full fine-tuning without tokenizer replacement}
\myTextMacro{\Fftwtk}{Full fine-tuning without tokenizer replacement}
\myTextMacro{\fleurs}{FLEURS}
\myTextMacro{\cv}{CommonVoice}
\myTextMacro{\whisper}{Whisper-large-v3}
\myTextMacro{\textonly}{text-only}
\myTextMacro{\Textonly}{Text-only}
\myTextMacro{\textaudio}{text--audio}
\myTextMacro{\Textaudio}{Text--audio}
\myTextMacro{\TextAudio}{Text--Audio}
\myTextMacro{\fishfood}{fish-food}
\myTextMacro{\model}{BuzzASR}

\newif\ifshowtodos
\showtodostrue     %

\ifshowtodos
    \newcommand{\todo}[1][]{%
        \sethlcolor{lime}
        \ifthenelse{\equal{#1}{}}{%
            \hl{TODO}}{%
            \hl{TODO: #1}%
        }
    }
    \newcommand{\tocite}[1][]{%
        \sethlcolor{yellow}
        \ifthenelse{\equal{#1}{}}{%
            \hl{CITE}}{%
            \hl{CITE: #1}%
        }
    }
    \newcommand{\alex}[1][]{%
        \sethlcolor{pink}
        \ifthenelse{\equal{#1}{}}{%
            \hl{ALEX}}{%
            \hl{ALEX: #1}%
        }
    }
    \newcommand{\catherine}[1][]{
        \sethlcolor{orange}
        \ifthenelse{\equal{#1}{}}{%
            \hl{CATHERINE}}{%
            \hl{CATHERINE: #1}%
        }
    }
    \newcommand{\aditya}[1][]{%
        \sethlcolor{red}
        \ifthenelse{\equal{#1}{}}{%
            \hl{ADITYA}}{%
            \hl{ADITYA: #1}%
        }
    }
\else
  \newcommand{\todo}[1][]{}
  \newcommand{\tocite}[1][]{}
  \newcommand{\alex}[1][]{}
  \newcommand{\catherine}[1][]{}
  \newcommand{\aditya}[1][]{}
\fi

\usepackage{longtable}
\usepackage{booktabs}
\usepackage{multirow}

\title{BuzzASR: A Swarm of 100+ Monolingual Speech Recognition Models}

\author{
  Shivam Singh$^1$ \quad
  Aditya Yadavalli$^1$ \quad
  Catherine Arnett$^2$ \quad
  Alex Warstadt$^1$ \vspace{0.5em}\\
  $^1$UC San Diego \quad $^2$EleutherAI \\
  \texttt{\{shs046, a1yadavalli, awarstadt\}@ucsd.edu}\hspace{2em}
  \texttt{catherine@eleuther.ai}
}
\begin{document}

\maketitle

\begin{abstract}
We introduce \model, a collection of language-specialized fine-tuned Whisper models adapted for automatic speech recognition (ASR) in 102 languages.
Large end-to-end Transformer-based ASR models such as Whisper \citep{radford2022whisper} have revolutionized ASR, but most prominent models are highly multilingual. 
As a result, these models often perform poorly on languages less well-represented in their training set.
While it has long been known that effective language adaptation can be achieved through simple fine-tuning on monolingual data, this strategy has only been applied to a small number of languages.
We massively scale up this simple approach to 102 languages covered in the FLEURS dataset, while also implementing a more complex language adaptation strategy that 
integrates monolingual tokenizer replacement and data augmentation using text-only fine-tuning.
\model models outperform \whisper on 77 out of 102 languages, reducing character error rates (CER) by a factor of over 2.8 on average.
Our models achieve state-of-the-art CER among open-source systems on 27
of 102 languages on the combined FLEURS and Common Voice test set. 
Our tokenizer replacement strategy yields an average 3.3{$\times$} improvement in compression rate (characters per token) over Whisper's multilingual BPE, with gains of up to 21.7{$\times$}.
We release all models, code, and detailed results:\\ \url{https://lemn-lab.github.io/buzz-asr}.
\end{abstract}

\section{Introduction}

\renewcommand{\topfraction}{0.95}
\renewcommand{\textfraction}{0.05}
\setcounter{topnumber}{4}
\begin{figure}[t!]
    \centering
    \begin{subfigure}[]{\columnwidth}
        \centering
        \includegraphics[width=\columnwidth]{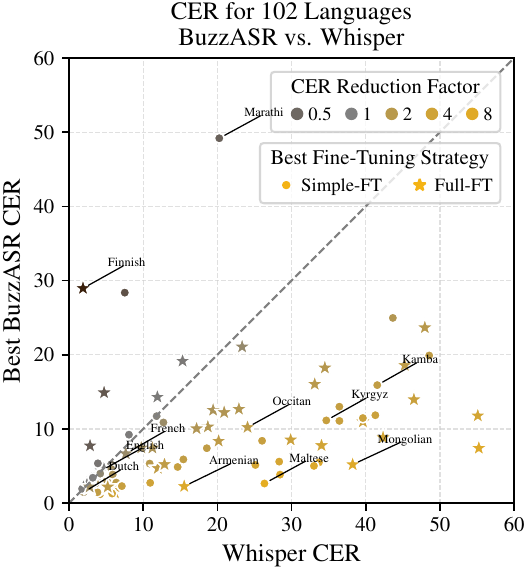}
        \label{fig:whisper_buzz_scatter}
    \end{subfigure}\vspace{-1em}
    \begin{subfigure}[]{\columnwidth}
        \centering
        \includegraphics[width=\columnwidth]{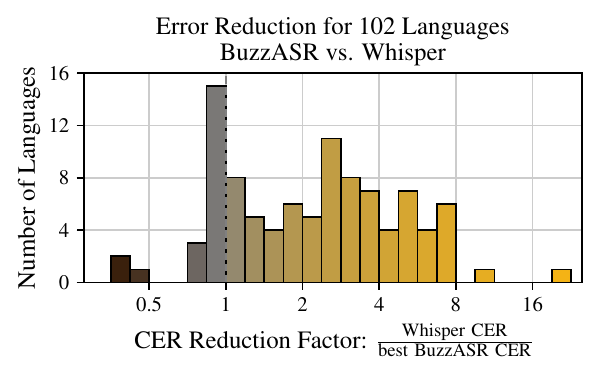}
        \label{fig:error_reduction_hist}
    \end{subfigure}
    \vspace{-2em}
    \caption{Comparison \model and Whisper on automatic speech recognition across 102 languages. For each language, we report only the best \model model. \emph{Some outliers are excluded for presentation.}
    }
    \label{fig:buzz_results}
\end{figure}

\begin{figure*}
    \centering
    \includegraphics[width=.94\textwidth]{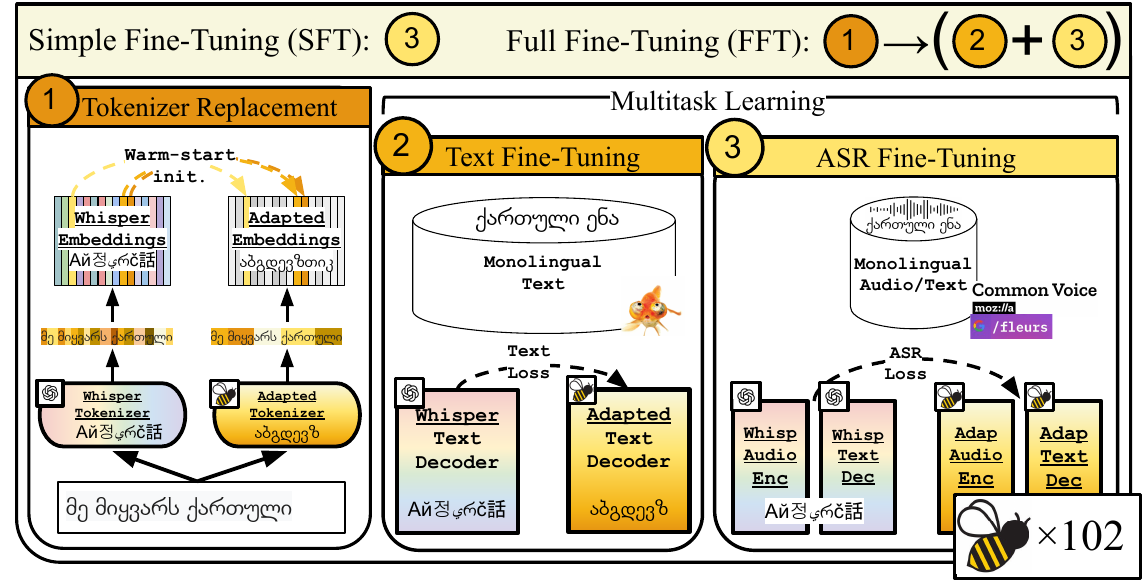}
    \caption{Overview of our two language adaptation approaches. \Vft is a minimal approach in which we fine-tune \whisper on monolingual speech recognition data. \Fft is a 3-stage pipeline involving \tkrep, \mtl on \textonly and speech recognition, and ending with speech recognition fine-tuning. }
    \label{fig:overview}
\end{figure*}
\renewcommand{\topfraction}{0.7}
\renewcommand{\textfraction}{0.2}
\setcounter{topnumber}{2}

Language technologies for the top 200 most widely spoken languages would serve approximately 90\% of the global population \citep{ethnologue200}. Currently, automatic speech recognition (ASR) models are highly inaccurate for all but a small fraction of those 200 languages.
This is not just due to a lack of data.
ASR datasets of $\geq$10 hours of speech recordings with text transcriptions exist for at least 1000 languages \citep{omnilingual2025}.
While this may not seem to be very much data, it can be surprisingly effective when applying domain adaptation to large pretrained multilingual ASR models.

Previous work was able to cut word error rates on low-resource languages by half compared to (at the time) state-of-the-art models, using less than 10 hours of monolingual data (\citealp[][\textit{inter alia}]{jimerson2018improving,liu2024exploration,ozyilmaz2025overcoming,imam-etal-2025-automatic}).
However, these contributions primarily applied this approach to a small number of domains or languages.
Thus, the reason why ASR models underperform for most widely spoken languages is not that we lack the ability to train such models; it is simply that the resources that exist have primarily been used to train either massively multilingual models or one-off language-specialized models.
Recently, in the text domain, it has been shown that small monolingual text generation models may outperform larger, massively multilingual models on linguistic tasks at the scale of hundreds of languages \citep{chang2026goldfish}. We take a similar approach to ASR and ask whether relatively small models dedicated to a single language can offer better performance, especially for lower-resource languages.

In this paper, we present a suite of 102 language-specific ASR models, which are fine-tuned models based on Whisper \citep{radford2022whisper}. 
For 77 of the 102 languages, the best of our fine-tuned models outperforms the original Whisper-Large-v3 on normalized CER, with the median language seeing CER cut by 2.18{$\times$} and the mean by 2.84{$\times$}. 
The median absolute CER drops from 15.5\% (Whisper zero-shot) to 7.6\% (best fine-tuned).
Among open-source ASR systems, our models achieve state-of-the-art CER on 27 of 102 languages on a combined evaluation drawn from FLEURS and Common Voice.
The gains are greatest where Whisper is weakest: in Whisper's 51 worst languages by CER, the median reduction is 3.45{$\times$}, and for ten languages the best fine-tuned model cuts CER by more than 6{$\times$}, led by Amharic, Maltese, and Sorani Kurdish.

While simple fine-tuning is effective, we also explore two improvements to the na\"ive approach:
First, we modify the Whisper tokenizer for each language. 
The original tokenizer was trained on an imbalanced multilingual dataset, leading to overly long sequence lengths---and therefore worse performance and inference latencies---in poorly represented languages or languages using non-Latin scripts.
Second, we fine-tune the Whisper text decoder on monolingual text data, in addition to fine-tuning both the audio encoder and text decoder on ASR data.
Many languages are poorly represented in Whisper's pretraining data, and the limited ASR data is insufficient to learn the target language's lexicon and grammar.
In many such cases, however, sufficient text data is available, allowing us to instill this language-specific knowledge without requiring large amounts of audio data.

The results of this work demonstrate that language-specific models can perform better and more efficiently, making use of relatively small resources. We release a suite of models for 102 languages, 27 of which are, to our knowledge, the state of the art for open-source models.

\section{Related Work}

\subsection{Low-Resource \& Multilingual ASR}
\label{sec:earlyasr}
Earlier work in low-resource ASR primarily focused on developing monolingual or limited multilingual systems for individual low-resource languages (\citealp[][\textit{inter alia}]{naing_myanmar2015,serbianasr2017, KipyatkovaRussianASR2016,bali2013, UpadhyayaHindiASR2017,AliMSAASR2014,luong-vu-2016-vietnamese,deka18_sltu}). 
Later, multilingual pretraining with little to no extra supervision was found to be effective \citep{watanabe17, toshniwal18}. 
Motivated by these findings, \citet{srivastava18_sltu} and \citet{diwan21_interspeech} released high-quality speech datasets and conducted shared tasks encouraging research in multilingual ASR. 
\citet{klejch21_interspeech} and \citet{mirishkar-etal-2021-investigation} achieved state-of-the-art results on the released shared task datasets, applying various established data-cleaning, augmentation, and multilingual techniques. 

In addition to large-scale data efforts, architectural innovations \cite{vaswanitransformers2017} and advancements in self-supervised learning have made it possible to build powerful scaled-up ASR models like Whisper that achieve state-of-the-art results on several benchmarks \cite{radford2022whisper}. 
Notably, wav2vec \cite{schneider19_interspeech} shows speech representations can be learned from unlabeled audio, leading to reduced dependency on labeled data. 
Many have scaled this paradigm up in terms of language coverage, number of training hours, and model size \cite{baevski2020wav2vec2, babu22_interspeech, zhang2023googleusmscalingautomatic, pratap2024mms}. 

At the time of release, Whisper achieved state-of-the-art performance across languages and diverse conditions, leading to widespread adoption \citep{radford2022whisper, olatunji2023afrispeech, bhogale2023vistaar, talafha2023arabic}. 
Whisper also motivated a range of adaptation strategies for new languages, domains, and tasks including fine-tuning \citep{bhogale2023vistaar, yadavalli2025prosodyconvey}, parameter-efficient fine-tuning \citep{kang2024whisperpeft, song2024lorawhisper}, prompt-tuning \citep{ma2024promptwhisper, yang2025promptwhisper}, multilingual or multitask fine-tuning \citep{bhogale2023vistaar, olatunji2023afrispeech}, and other methods \citep{zhao2026lorsmerging, juvekar2026vividhasr}. 
Given the broad adoption and the growing body of work on adapting Whisper, we experiment with Whisper models in this study.

\subsection{ASR Datasets}
\label{sec:datasets}

Recent efforts in creating and publicly releasing large multilingual ASR corpora have enabled the scaling of multilingual ASR models \cite{pratap20_interspeech, wang-etal-2021-voxpopuli, li24s_msr}. 
However, many of these datasets primarily focus on high-resource European languages. 
To address this limitation, substantial efforts have been made to collect speech corpora for other languages and regions. 
One such early effort is BABEL \cite{gales14_babel}, a multilingual conversational telephone speech corpus covering 17 low-resource languages. 
Common Voice \cite{ardila-etal-2020-common} is a continuously evolving large-scale crowdsourced effort maintained by Mozilla to collect speech data across a large number of languages. 
FLEURS \cite{conneau_fleurs} is an $n$-way parallel speech dataset covering 102 languages and is adapted from the FLORES-101 \cite{goyal-etal-2022-flores} machine translation benchmark. 
\citet{chen-etal-2024-xeus} release an un-transcribed multilingual speech dataset with coverage spanning more than 4000 languages. 
Most recently, Omnilingual released transcribed data for 348 traditionally under-resourced languages \citep{omnilingual2025}.

Several region-specific initiatives have also focused on improving speech resources for underrepresented languages. %
Project Vaani \cite{pulikodan2026vaanicapturinglanguagelandscape} released more than 30,000 hours of culturally relevant speech data spanning over 100 languages in the Indian subcontinent while ensuring regional diversity. Shrutilipi \cite{bhogale_shrutilipi} released more than 6,000 hours of transcribed Indic speech data mined from in-the-wild sources. Similarly, African Next Voices \cite{marivate2026swivurisosouthafricanvoices}, NaijaVoices \cite{emezue25_interspeech}, and WAXAL \citep{diack2026waxal} provide high-quality culturally rich speech datasets for several African languages. In this work, we use Common Voice \cite{ardila-etal-2020-common} and FLEURS \cite{conneau_fleurs} because they provide high-quality multilingual speech data with broad language coverage, making them well-suited for our work.

\subsection{Curse of Multilinguality}

While joint multilingual training can lead to impressive results due to crosslingual transfer for ASR \citep{pratap2024mms, omnilingual2025}, multilingual pretraining may also be limited by the ``curse of multilinguality" \citep{conneau-etal-2020-unsupervised, mei2026zipperlora, zhao2026lorsmerging}. This refers to the phenomenon where training on a very large number of language leads to degraded performance for many of the languages. 
This has been studied for text generation models, and has been found to be driven by limited model capacity and negative interference from unrelated languages \citep{chang-etal-2024-multilinguality}.

\section{Adaptation strategies}
\label{sec:adaptation_strategies}

In this paper, we hope to challenge the reliance on the one-model-to-rule-them-all paradigm and advocate for language- and context-specific training to maximize performance for each language.
We consider two approaches to adapting pretrained Whisper to a target language: 
\newterm{\Vft} entails fine-tuning the entire Whisper model on aligned monolingual speech--text data using the ASR objective.
\newterm{\Fft} is a more complex pipeline in which we replace Whisper's multilingual tokenizer with a language-specialized one and fine-tune on a mixture of monolingual text-only and speech--text data.

Referring to \Cref{fig:overview}, we introduce the basic building blocks of these pipelines below:
\newterm{\tkrep}, \newterm{\asrft}, and \newterm{\textft}. Additionally, \newterm{\mtl} combines \textft and \asrft. 

\subsection{Tokenizer Replacement}
\label{sec:tk_rep}

Whisper uses a byte-level byte-pair encoding (BPE) tokenizer \citep{sennrich-etal-2016-neural,radford2019language} with a vocabulary of 51{,}865 tokens: 50{,}257 BPE merges and 1{,}608 special tokens, including 99 language tokens. 
The Whisper tokenizer was trained on a corpus heavily skewed towards English  \citep{radford2022whisper}. As a result, the tokenizer is well-adapted only to the most highly represented languages, and is especially poorly suited to languages with non-Latin and unique scripts. Replacing the vocabulary allows the models to maximally and efficiently benefit from the language-specific adaptation (\citealp[][\textit{inter alia}]{minixhofer-etal-2022-wechsel}).

To perform \tkrep for a target language, we begin by training a new BPE tokenizer on a monolingual text corpus.
While we could simply randomly initialize a new embedding matrix for the new vocabulary, this approach fails to leverage existing structure in the pretrained model. 
Instead, we adopt a \newterm{warm-start initialization} of the embedding matrix, resembling \citet{gee-etal-2022-fast}.
First, embeddings for any entries appearing in both tokenizers' vocabularies are initialized using the original Whisper embedding.
Second, for any entries appearing only in the new tokenizer's vocabulary, we tokenize that entry according to the original tokenizer, and initialize the new embedding as the average of the original tokenizer's embeddings.

For example, the German compound \emph{Geschwindigkeit} (``speed'') receives its own dedicated entry in our per-language German BPE. Whisper's original tokenizer, lacking a single merge for this compound, decomposes it into four subword tokens: \texttt{G} + \texttt{esch} + \texttt{wind} + \texttt{igkeit}. 
The new \emph{Geschwindigkeit} embedding is initialized as the mean of those four subword embeddings, preserving whatever lexical information Whisper had already encoded at the level of static embeddings.

We apply the identical warm-start procedure to the output (unembedding) projection, initializing each row from (the average of) the corresponding Whisper output embeddings, so that both the input and output representations of a new token inherit Whisper's encoded information. 
Throughout, we keep the token-to-index mapping consistent between the embedding and unembedding matrices.

\subsection{ASR Fine-Tuning} 
\label{sec:asr_ft}
During \asrft, the entire Whisper encoder--decoder model is fine-tuned on aligned speech--text data using an ASR objective. 
Our \vft pipeline consists only of this step.

\subsection{Text Fine-Tuning}
\label{sec:text_ft}
During \textft, the Whisper text decoder is fine-tuned on text-only data using the autoregressive language modeling objective (next-token prediction).
As Whisper's decoder incorporates cross-attention from the encoder embeddings, we pass dummy inputs to the encoder in the form of random embeddings and freeze the encoder.

Text fine-tuning is primarily useful insofar as text-only data is more abundant than speech--text data for the target language.
Generally, there are several orders of magnitude difference in dataset size. 
For some languages, the only ASR data available is from FLEURS, comprising only about 40,000 English words, while we have at least 1M tokens of text data for each language---and in many cases much more data. 
While in most cases we would prefer \asrft given sufficient data, fortunately, there are two areas in which we speculate that training the decoder alone on text data is sufficient:

First, the syntactic priors of the model can conceivably be improved through \textft. 
The decoder alone is responsible for outputting grammatical text, while the encoder may specialize in acoustic processing and phonological representations. 
\citet{chang2026goldfish} showed that even with less than 1GB of text data, small models can learn syntactic information as well as much larger models trained on orders of magnitude more data.
Not only is the syntax largely dependent on the text alone, but syntax differs substantially across languages. 
Text-generation models exhibit worse cross-lingual transfer when they have less similar syntax \citep{chang-etal-2024-multilinguality}. 
If syntax is more language-specific and harder to learn, we reason that the decoder stands to benefit from relatively abundant text-only input. 

Second, if we are performing tokenizer replacement, the model must adapt to the new vocabulary.
As this is likely to require extensive training, and the tokenizer is used only by the decoder, text-only inputs are particularly valuable at this stage to prevent tokens in the new vocabulary from being under-trained.

\subsection{Multitask Learning}
\label{sec:mtl}
\Mtl is a single fine-tuning stage in which each minibatch contains a mixture of ASR examples (aligned speech--text pairs) and text-only examples, with the fraction of ASR examples controlled by a hyperparameter $\alpha \in (0, 1]$, the \emph{ASR proportion}. 
Each example in the batch contributes its own loss term: the ASR loss for aligned pairs, computed against the full encoder--decoder output; and the language-modeling loss for text-only examples, computed by passing random embeddings as encoder input and freezing the encoder, as described in \Cref{sec:text_ft}. 
Setting $\alpha = 1.0$ recovers pure \asrft; setting $\alpha = 0$ recovers pure \textft.

\section{Experiments}
\label{sec:experiments}

We study and compare two fine-tuning strategies (\vft and \fft) to adapt a multilingual model to 102 target languages. 
All our experiments are done using the 1.55B parameter \whisper model. 
For each fine-tuning strategy and language, we conduct three fine-tuning runs with different hyperparameter configurations, resulting in 612 fine-tuned models.
We report our hyperparameters in  \Cref{app:hps}.

\subsection{Data and Languages}

\paragraph{\TextAudio Data}

We draw aligned \textaudio data from two sources: \fleurs \citep{conneau_fleurs} and \cv v25 \citep{ardila-etal-2020-common}. 
We limit all our experiments and models to the 102 languages represented in \fleurs.
Due to compute limitations, we restrict speech--text data to 100 hours (in English, this is approximately 1 million words) for higher-resource languages.

\fleurs is an $n$-way parallel corpus consisting of English Wikipedia data expert-translated into 102 languages with audio recordings by 3 native speakers.
The training set contains approximately 10 hours per language, and the development and test splits contain on the order of 200--1000 samples per language depending on language coverage.

\cv is a crowdsourced dataset with variable coverage of almost 300 languages.
We use only data from the 77 languages also in \fleurs. 
Due to vast differences in data quantities per language and computational constraints, we cap the \cv training data at 90 hours per language.
To equalize compute across languages with very large test sets (e.g., English, Russian, Mandarin), we cap each language's test data at 2{,}000 utterances. 
This cap is reached for $36$ of the $77$ languages that are included in \cv.\footnote{Languages whose CV-test split is below 2{,}000 are evaluated on all available samples.} 
The remaining $25$ out of 102 languages did not have any \cv data, so we use only \fleurs data. 

\fleurs provides normalized transcripts, which we use as-is. For \cv, we apply only lightweight text normalization: lowercasing, punctuation removal with a shared regular expression, and whitespace normalization. We keep the preprocessing minimal so as to avoid altering transcript content beyond surface formatting.

For all datasets, audio is resampled to 16~kHz, and model inputs are limited to at most 30 seconds of audio.
We use the provided test splits for both datasets for evaluation.

\paragraph{Text Data} 
We sample 500k lines from the \fishfood corpus \citep{chang2026goldfish}. O
nly one \fleurs language (Kamba) had no data in \fishfood, so we use the \fleurs training transcripts as \textonly data. 
Text-only training corpora are NFC-normalized,\footnote{This is a standard unicode normalization that combines separate diacritics and base characters into a single, precomposed character.} lowercased, punctuation-stripped, filtered to require $\geq 70\%$ characters in the language's primary script, and deduplicated based on exact string match.

\paragraph{17-Language Subset}
For some supplementary experiments, we study a subset of the 102 FLEURS languages, selected heuristically to represent a wide variety of language families, scripts, and resource levels. 
These languages, along with key metadata, are listed in \Cref{tab:core_languages}.

\subsection{Tokenizer Replacement}
For each language, we conduct \tkrep as described in \Cref{sec:tk_rep}.
We train a 51{,}865-token byte-level BPE tokenizer per language, matching Whisper's original vocabulary size. 
This allows us to apply warm-start initialization without resizing the special-token block or the decoder's unembedding head. 
We use the same pre-tokenizer as Whisper. 
Each tokenizer is trained on the same per-language \fishfood corpus used for \textft, so the tokenizer's vocabulary matches the distribution the decoder will be fine-tuned on.

\subsection{Multitask Learning}
For \mtl experiments, we interleave text-only and ASR samples within a single fine-tuning stage. 
At each optimization step, we sample either an aligned speech-text sample with probability $\alpha$ or a text-only sample. 
In a supplementary experiment, we tested different mixing ratios of text-only and ASR samples (10\%, 20\%, and 50\% text data; \Cref{app:mtl_mixing}) with a subset of languages, and found that 50\% text data was optimal. 
Therefore, this is the mix that we used.

\subsection{Experimental Configurations}
\label{sec:hyperparams}

For each of the 102 target languages and each fine-tuning strategy, we train a small set of learning-rate configurations and select, per language, the one with the lowest word error rate on the FLEURS development split. 
These configurations (\textbf{cfg~A}, \textbf{cfg~B}, and \textbf{cfg~C}) set absolute learning rates for \vft.
For \fft, the configurations set the embedding- and decoder-learning-rate multipliers on a fixed
base rate. 
We determined this set of configurations heuristically following a hyperparameter search with a subset of 17 languages (\Cref{app:hps}).

\paragraph{Optimization.}
All runs use AdamW \cite{loshchilov2018adamw} with a per-device batch size of 2 and patience-based early
stopping on FLEURS-validation WER, evaluated every 100 steps (patience 3 for \vft, 5 for \fft), for at most 6 epochs. 
\Vft uses a 150-step warmup and weight decay ${\approx}4\times10^{-5}$. 
\Fft uses a 200-step warmup and weight decay $0.01$ (per-configuration values in \Cref{tab:configs}). 
We compute cross-entropy loss over the label tokens, masking padding positions with $-100$.
For text-only batches in \fft, the encoder receives random embeddings and is frozen during the backward pass, so gradients flow only through the decoder.

\subsection{Evaluation}
Our primary evaluation metric is normalized character error rate (CER). 
We report both CER and WER in \Cref{app:full_results}, but focus our discussion on CER since word length varies widely across languages and makes WER less comparable.
All reported numbers are on the combined FLEURS\,+\,Common Voice test split; for the 25 languages not covered by Common Voice, we report FLEURS results only. 
CER and WER are computed with \texttt{jiwer}.\footnote{\url{https://github.com/jitsi/jiwer}}

We also report Whisper zero-shot as a baseline. Of the 102 target languages,
83 are supported by Whisper, which we decode using the corresponding Whisper
language token. 
The remaining 19 languages are not supported by Whisper; for
these we prompt the model with the most closely related supported language
(e.g., Asturian with Spanish, Kyrgyz with Kazakh, Oriya with Bengali). 
The full
mapping is given in \Cref{app:proxy}. 
All systems including the Whisper
zero-shot baseline are decoded identically with greedy search
(\texttt{num\_beams=1}, \texttt{no\_repeat\_ngram\_size=3},
\texttt{repetition\_penalty=1.2}) and no external language-model fusion.

\subsection{Baselines}

We compare our methods to the original \whisper models and several recently released multilingual ASR models: Omnilingual 1B and 7B \citep{omnilingual2025}; Massively Multilingual Speech (MMS, \citealp{pratap2024mms}); Qwen3-ASR 1.7B \citep{Qwen3-ASR}; and Cohere Transcribe 2B \citep{julian_mack_2026}. 
We report raw CER/WER for Qwen3-ASR and Cohere Transcribe, which do not provide normalized scores.
We evaluate on all 102 languages, although not all languages are supported by every model.

\section{Results}

\begin{figure}
    \centering
    \includegraphics[width=\linewidth]{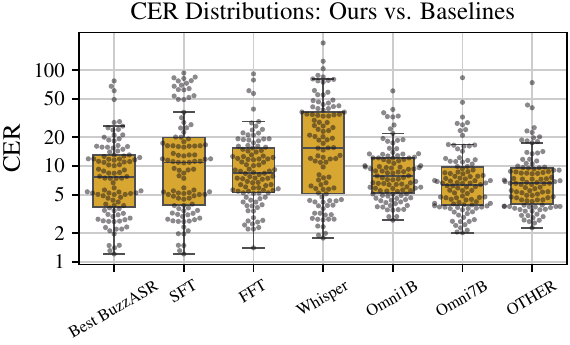}
    \caption{CERs for 102 languages for Whisper, \model models, Omnilingual models, and all other baseline models (only the best per language is shown).}
    \label{fig:baseline_swarmplot}
\end{figure}

\begin{table}[t]
\centering
\small
\setlength{\tabcolsep}{4pt}
\resizebox{\columnwidth}{!}{%
\begin{tabular}{lcccccc}
\toprule
 & \multicolumn{2}{c}{\textbf{Simple-FT}} & \multicolumn{2}{c}{\textbf{Full-FT}} & \multicolumn{2}{c}{\textbf{Best BuzzASR}} \\
\cmidrule(lr){2-3} \cmidrule(lr){4-5} \cmidrule(lr){6-7}
Baseline & CER & WER & CER & WER & CER & WER \\
\midrule
Whisper-ZS & 65 & 65 & 69 & 72 & 77 & 78 \\
Omni-1B & 52 & 60 & 44 & 65 & 60 & 76 \\
Omni-7B & 36 & 40 & 31 & 43 & 42 & 53 \\
MMS-1B & 41 & 54 & 39 & 58 & 53 & 70 \\
Qwen3-ASR & 91 & 99 & 91 & 99 & 94 & 100 \\
Cohere-T & 99 & 102 & 93 & 100 & 99 & 102 \\
\midrule
All & 19 & 22 & 19 & 34 & 27 & 39 \\
\bottomrule
\end{tabular}
}
\caption{Win rates against external baselines on the combined FLEURS+CV25 test set. Each cell is the number of languages (of 102) where the given \model strategy achieves lower CER/WER than the baseline. Best is the better of Simple-FT and Full-FT per language.}
\label{tab:win_rate}
\end{table}

In this section, we compare the performance of \model to Whisper and other pretrained ASR models, as well as reporting the efficacy of tokenizer adaptation.
The complete CER and WER rates for \model and other pretrained models on all 102 languages are reported in \Cref{tab:test_all_languages_cer} and \Cref{tab:test_all_languages_wer}, respectively. 
Complete tokenizer performance metrics for the adapted tokenizers and the original Whisper tokenizer are reported in \Cref{tab:per_lang_tokenizer}.

\subsection{ASR Results}

As the \model models are fine-tuned from \whisper, we first compare the performance between the two. \Cref{fig:buzz_results} compares CER on the combined test split for \whisper vs.~\model.
We report the best performance among our two fine-tuning strategies. 
We find that the \model models outperform Whisper on 77 languages, reducing CER by a factor of 2.8 on average.
The scatter plot (\Cref{fig:buzz_results}; top) shows that not only do most points fall below the diagonal, indicating that \model generally improves on \whisper, but most points are well below the diagonal, showing substantial improvement.
Improvements are especially pronounced for languages where Whisper CER is high.
The majority of languages where we see little or no improvement are languages where Whisper CER is below 10, including English, Russian, French, Dutch, and Swedish.
This suggests that fine-tuning is most valuable for languages under-served by Whisper's original training distribution, while for high-resource languages, Whisper can remain competitive. 
For a small set of languages, such as Finnish and Marathi, we find that fine-tuning leads to substantially \emph{worse} performance.
We infer that fine-tuning can sometimes be unstable, and leave open the possibility that additional fine-tuning runs could still benefit these languages.

To determine the magnitude of these improvements, we calculate the CER reduction factor for each language, which we define as the ratio between Whisper zero-shot CER and the lower CER achieved by either \vft or \fft.
For example, a value of 2 indicates that \model achieves half the CER of \whisper, and any value greater than 1 indicates an improvement.
The distribution of CER reduction factors is visualized in \Cref{fig:buzz_results} (bottom).
Across 102 languages, the median reduction factor is 2.2, meaning that the best fine-tuned system more than halves the CER for the median language. 
On the other hand, the mean reduction factor is much higher at 2.8, reflecting large gains for several languages with very poor zero-shot Whisper performance. 
In the strongest cases, such as Amharic, Armenian, Maltese, Assamese, Uzbek, and Kabuverdianu, fine-tuning reduces CER by more than 6{$\times$} relative to Whisper zero-shot.

In \Cref{tab:win_rate}, we show the win rates for \model models vs.~all external baselines.
We find that \vft improves the CER for 65 languages, while \fft is better for 69 languages. 
This suggests that neither fine-tuning method is optimal for all languages, but instead \fft is more suitable for languages where the base model has a high error rate and \vft is more appropriate for languages the model already makes fewer errors on. 

Next, we compare to other open-weight models.
Importantly, \model beats all baselines on CER for 27 languages reaching SOTA performance.%
\footnote{\model outperforms all baselines on 39 languages when using WER as an evaluation metric.}
Omni-7B is the strongest baseline, with our best model outperforming it on only 42 languages, using CER. 
However, against the more comparably sized Omni-1B, \model performs better for 60 languages.
The best \model model per language outperforms Qwen3-ASR and Cohere Transcribe on nearly every language (94 and 99, respectively).
The full distribution of CERs for most models are visualized in \Cref{fig:baseline_swarmplot}.

\subsection{Tokenizer-quality metrics}
\label{sec:tokenizer_results}

\begin{figure}[t]
    \centering
    \begin{subfigure}[]{\columnwidth}
        \centering
        \includegraphics[width=\columnwidth]{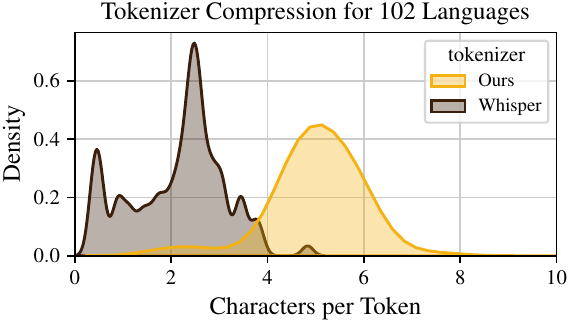}
        \label{fig:tokenizer_kde}
    \end{subfigure}\vspace{-1em}
    \begin{subfigure}[]{\columnwidth}
        \centering
        \includegraphics[width=\columnwidth]{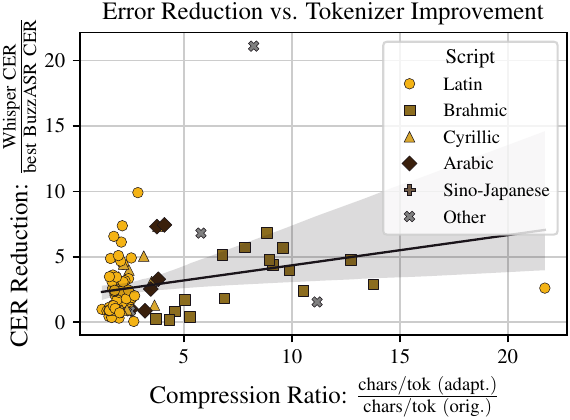}
        \label{fig:compression_cer_reduction}
    \end{subfigure}
    \vspace{-2em}
    \caption{\textbf{Top}: Distribution of tokenizer compression rates for all languages with original and adapted tokenizers. \textbf{Bottom:} Relationship between improvement in CER between Whisper and \fft by writing system and change in compression rate after tokenizer adaptation.}
    \label{fig:tokenizer_results}
\end{figure}

Next, we analyze the impact of tokenizer adaptation. \Cref{fig:tokenizer_results} (top) compares the compression rate, measured as characters per token, of our tokenizer and Whisper's tokenizer across all the evaluated languages.
The density plot shows that our models have higher compression rates for the vast majority of languages. 
The median compression rate increases from 2.38 characters per token for Whisper to 5.04 for our tokenizers, resulting in a median gain of 2.18{$\times$}. 
The effect is especially pronounced for non-Latin scripts, where Whisper's tokenization is highly fragmented the median compression gain is 3.80{$\times$} for non-Latin scripts versus 1.93{$\times$} for Latin scripts.
This reduction in fragmentation shortens sequence lengths, thus making multilingual speech recognition more efficient. 
We refer the readers to \Cref{tab:latency-by-script}, which shows the script-wise improvement in ASR latency due to our improved tokenizers. 
Furthermore, the distribution of compression rates across languages appears roughly normal, which suggests that the compression rates are more consistent cross-linguistically. 

We also examine how the improvements in tokenization efficiency are associated with downstream ASR gains using \Cref{fig:tokenizer_results} (bottom). 
The $x$-axis shows the compression rate ratio between our tokenizer and Whisper's tokenizer, while the $y$-axis shows the CER reduction factor obtained by the \model model with \fft relative to Whisper zero-shot. 
The regression line shows a general positive relationship.
This indicates that languages for which our tokenizer provides larger compression gains also tend to show larger CER reductions after fine-tuning. 
However, this trend is moderate rather than deterministic, indicating that tokenization is only one of the many factors affecting the ASR performance. 
The effect is especially pronounced for non-Latin script languages---they have a higher median compression ratio than Latin-script languages and also tend to obtain larger CER reductions. 
We report additional tokenizer metrics, such as UTF-8 coverage, vocab utilization, and boundary crossing, in \Cref{app:tokenizer_per_lang}.

\subsection{Supplementary Results}

In the Appendix, we report several supplementary experiments.
We briefly summarize those findings below.

\paragraph{For \mtl, higher proportions of ASR training are better.}
In \Cref{app:mtl_mixing}, we explore variants of the \mtl training hyperparameters.
Among the settings we compare, 50\% ASR data  is optimal, and increasing the amount of text-only data yields modest improvements.

\paragraph{Warm-start initialization improves tokenizer adaptation.}
In \Cref{app:tokenizer_strategy}, we compare alternative strategies for tokenizer adaptation, including a tokenizer replacement strategy where we randomly initialize new language-specific tokens and a tokenizer adaptation strategy where we append new tokens to the end of the original tokenizer merge list.
The warm-start tokenizer replacement strategy used in our main \fft experiments performs best.

\paragraph{Out-of-domain evaluation hurts performance slightly.}
In \Cref{app:ood}, we test out-of-domain generalization of models trained on FLEURS alone.
We find that out-of-domain evaluation hurts performance slightly.
This suggests that some (but probably not all) of the performance gains we observe over baseline models are due to better domain alignment of our models' train and test data, rather than language-specific fine-tuning.

\paragraph{Multilingual fine-tuning harms performance slightly.}

In \Cref{app:multilingual}, we compare bilingual fine-tuning runs to monolingual runs.
For nearly all language pairs (including related and unrelated language pairs), we find that monolingual fine-tuning is more effective.

\paragraph{\Mtl is not helpful without tokenizer replacement.}

In \Cref{app:mtl_tk_ablation}, we disentangle the effects of tokenizer replacement and \mtl, which are conflated in our \fft experiments.
We find that \fft is more effective than \mtl without tokenizer replacement.

\section{Discussion and Conclusion}

\model models provide SOTA or near-SOTA automatic speech recognition for dozens of languages.
Given their Whisper-based architecture, \model models can be easily deployed in any pipeline already implemented for Whisper where strong monolingual performance is needed on one of our covered languages.
While \model models are outperformed by Omni-7B on the majority of languages, we note that \model generally outperforms Omni-1B, which has a more comparable number of parameters.
Thus, in cases where Omni-7B is too large to deploy, \model offers competitive performance for the majority of languages we cover.

In addition to providing a new open-source model suite, our experiments demonstrate the efficacy of language-specialized fine-tuning at scale.
Given the benefits of language-specialized models to the global community, our work motivates similar approaches with other ASR model architectures, as well as  with text models and multimodal models.

Finally, our tokenizer replacement approach reveals the extent to which multilingual tokenizers under-deliver, particularly for non-Latin scripts.
We show that language-adapted tokenizers not only improve ASR performance disproportionately in cases where the original tokenizer was ill-suited, they also decrease inference latency and compute by reducing sequence lengths.

\paragraph{Future Work}
There are significant opportunities for future improvements to \model.
First, we believe through further hyperparameter-tuning for vocabulary adaptation and multi-task learning we could achieve better performance. 
We focused on a limited set of design choices to optimize, and most significantly improved performance. 
By further tuning the multitask learning protocol, we could better make use of abundant text-only data.

Second, there is a wide array of established optimizations for ASR that we do not yet exploit.
These include data augmentation through the addition of noise or text-to-speech, interpolation with monolingual text-only LMs, and training simultaneously on closely related languages.

Finally, we focus on a relatively small set of languages. 
As we note above, there is sufficient fine-tuning data for roughly 1000 languages. 
We hope to expand our suite of \model models, especially to include under-represented and under-studied languages.
As ASR can be used to increase text data for training language models, creating better tools for these languages can lead to improvements in other language technologies.

\section*{Limitations}

In this paper, we only conduct fine-tuning experiments on Whisper. 
However, Omni-7B still shows state-of-the-art performance for most languages.
We anticipate that fine-tuning Omnilingual would result in even better performance. 
We focused on Whisper because it is very widely used and integrated into many existing pipelines. 
Replacing Whisper with one of our models would require minimal adjustments for users. 
Furthermore, \whisper is significantly smaller than Omni-7B, with only 1.6B parameters. 
Focusing on smaller models results in more efficient and economical models, and we note that \model models outperform the similarly sized Omni-1B in the vast majority of cases.
While we aimed to produce performant models, the main goal of the paper was to demonstrate the efficacy of our fine-tuning approach for models optimized for a single language. 
Future work could explore using other models as starting points.

We used only a small fraction of the existing ASR data. 
We limited the data we use to 100 hours of data, primarily due to computational resource limitations.  
For the languages we fine-tuned models for in this paper, we could achieve even higher performance by using more of the available data. 
There are also many languages which we did not train models for, but for which there is a significant amount of data. 
We hope this approach can be applied to a wider set of languages. 

We use the test splits from our training datasets (FLEURS and Common Voice) for evaluation. 
Due to data limitations, we do not evaluate performance across different domains or dialects, as such datasets are not available for many of the languages we consider in this paper.
A supplementary analysis \Cref{app:ood} reveals that out-of-domain generalization may be a challenge for adapted ASR models resembling the \model releases.

We also note that monolingual models may provide better performance in controlled evaluation, but might not be appropriate for all applications. 
Monolingual models are best when one already knows the language that is being used, which may limit their utility. 
If coverage of multiple languages is required, deploying multiple monolingual models will require substantially more compute resources than deploying a single multilingual model.
This is especially true given we perform full fine-tuning, rather than parameter-efficient fine-tuning, which improves memory efficiency in exchange for some losses in performance on ASR \citep{liu2024exploration}.
Additionally, multilingual models may better be able to handle code switching or multi-dialectal speech, though this may require additional specialized training or fine-tuning.

\section*{Acknowledgments} %
This work used resources available through the National Research Platform (NRP) at the University of California, San Diego \citep{nautilus}. 
NRP has been developed, and is supported in part, by funding from National Science Foundation, from awards 1730158, 1540112, 1541349, 1826967, 2112167, 2100237, and 2120019, as well as additional funding from community partners. 
We also thank CoreWeave for providing compute for this project.

\bibliography{custom}

\clearpage

\appendix

\section{Unsupported Whisper Languages} \label{app:proxy}
Out of the 102 languages we experiment with, 19 of them are not natively supported by Whisper. For such languages, we prompt the model with the language token that is most closely related to the target language. These languages are Asturian, Cebuano, Sorani Kurdish, Fulah, Irish, Igbo, Kamba, Kabuverdianu, Kyrgyz, Luganda, Luo, Northern Sotho, Nyanja, Oromo, Oriya, Umbundu, Wolof, Xhosa, and Zulu.

\begin{table*}[ht]
\centering
\small
\begin{threeparttable}
\begin{tabular}{@{}llllll@{}}
\toprule
\textbf{Language} & \textbf{Code} & \textbf{Script} & \textbf{Family} & \textbf{Region} & \textbf{Speakers} \\
\midrule
Arabic     & ar & Arabic             & Afro-Asiatic (Semitic)\tnote{a} & MENA              & 411M \\
Cantonese  & --\tnote{b} & Chinese (Han) & Sino-Tibetan            & S.\ China / Hong Kong   & 85M \\
Catalan    & ca & Latin              & Indo-European (Romance)   & NE Spain / Andorra      & 4.1M \\
Galician   & gl & Latin              & Indo-European (Romance)   & NW Spain (Galicia)      & 2.4M \\
Georgian   & ka & Georgian           & Kartvelian                & South Caucasus         & 3.76M  \\
German     & de & Latin              & Indo-European (Germanic)  & Central Europe         & 95M  \\
Italian    & it & Latin              & Indo-European (Romance)   & Southern Europe        & 65M  \\
Latvian    & lv & Latin              & Indo-European (Baltic)    & Baltic region          & 1.5M  \\
Pashto     & ps & Perso-Arabic       & Indo-European (Iranian)   & Afghanistan / Pakistan  & 51M  \\
Polish     & pl & Latin              & Indo-European (Slavic)    & Central Europe         & 40M  \\
Russian    & ru & Cyrillic           & Indo-European (Slavic)    & E.\ Europe / N.\ Asia   & 145M  \\
Spanish    & es & Latin              & Indo-European (Romance)   & Iberia / Americas       & 519M  \\
Swahili    & sw & Latin              & Niger--Congo (Bantu)      & East Africa            & 5.3M \tnote{c} \\
Tajik      & tg & Cyrillic           & Indo-European (Iranian)   & Central Asia           & 10.5M  \\
Tamil      & ta & Tamil              & Dravidian                 & S.\ India / Sri Lanka   & 79M  \\
Ukrainian  & uk & Cyrillic           & Indo-European (Slavic)    & Eastern Europe         & 32M  \\
Uzbek      & uz & Latin              & Turkic                    & Central Asia           & 34M  \\
\bottomrule
\end{tabular}
\caption{Overview of 17 sample languages, with ISO 639-1 codes, writing systems, language families, primary geographic regions, and approximate number of first-language (L1) speakers, based on Wikipedia data as of 2026 (mainly citing Ethnologue).}
\label{tab:core_languages}
\begin{tablenotes}
\footnotesize
\item[a] Cantonese lacks an ISO 639-1 code, being covered by the Chinese macrolanguage code \texttt{zh}. Its ISO 639-3 code is \texttt{yue}.
\item[b] This figure reflects the combined L1 total across all varieties under the Arabic macrolanguage as classified by Ethnologue/ISO.
\item[c] Swahili L1 speakers are vastly outnumbered by L2 speakers (approx.\ 92M L2 vs.\ 5.3M L1).
\end{tablenotes}
\end{threeparttable}
\end{table*}

\section{Hyperparameters}
\label{app:hps}

\subsection{Hyperparameters for \mtl}

For each language and each fine-tuning strategy we train models with three configurations, A, B, C (\Cref{tab:configs}).
We use two random seeds (42, 1337) for simple fine-tuning and three (42, 1337, 2024) for full fine-tuning.

For \vft, we use an absolute decoder learning rate and set the encoder rate to $0.3\times$ the decoder rate, so as to preserve Whisper's pretrained acoustic features.

We apply patience-based early stopping on FLEURS development throughout. 
For \fft, we use a patience of 5 and evaluate every 100 steps. 
For \vft, we use a patience of 3 and evaluate every 100 steps and select checkpoints by validation WER. 
For \fft, we select checkpoints by validation WER for high-coverage languages but use validation loss for a small set of low-coverage languages where the WER is noisy.\footnote{\Fft loss-based selection is used for: Tamil, Latvian, Uzbek, Cantonese, Pashto, Swahili, Tajik, Georgian, and Ukrainian. This split was determined empirically.}

We use 8$\times$A100-80GB for all experiments.

\begin{table*}[t]
\centering
\small
\setlength{\tabcolsep}{3pt}
\begin{tabular}{l r r r r l r r r}
\toprule
\bf Cfg & \bf dec LR & \bf enc LR & \bf embed LR & \bf ga & \bf sched. & \bf decay & \bf warmup & \bf wd \\
\midrule
\multicolumn{8}{@{}l}{\textbf{SFT}} \\
A & $6.48{\times}10^{-6}$ & $0.3{\times}\textnormal{dec LR}$ & -- & 16 & cosine & 0.95 & 150 & $4.35{\times}10^{-5}$ \\
B & $2.00{\times}10^{-5}$ & $0.3{\times}\textnormal{dec LR}$ & -- & 32 & cosine & 0.95 & 150 & $4.00{\times}10^{-5}$ \\
C & $3.24{\times}10^{-6}$ & $0.3{\times}\textnormal{dec LR}$ & -- & 16 & cosine & 0.95 & 150 & $4.35{\times}10^{-5}$ \\
\midrule
\multicolumn{8}{@{}l}{\textbf{FFT}} \\
A & $3.0{\times}10^{-5}$ & $3\times 10^{-6}$ & $3.0{\times}10^{-5}$ & 16 & cosine-delay & -- & 200 & 0.01 \\
B & $1.0{\times}10^{-5}$ & $3\times 10^{-6}$ & $3.0{\times}10^{-5}$ & 16 & cosine-delay & -- & 200 & 0.01 \\
C & $2.0{\times}10^{-5}$ & $3\times 10^{-6}$ & $2.0{\times}10^{-5}$ & 16 & cosine-delay & -- & 200 & 0.01 \\
\bottomrule
\end{tabular}
\caption{Fine-tuning configurations, fully specified. \textbf{dec/embed LR}: decoder/embedding learning rate; \textbf{ga}: gradient accumulation; \textbf{sched.}: LR schedule (cosine-delay = cosine with a 1000-step delay, 8000-step period); \textbf{decay}: scheduler decay rate; \textbf{wd}: weight decay. All runs use AdamW and micro-batch size~2 (effective batch $=2{\times}$ga). SFT uses a single learning rate (no separate embedding LR); FFT scales a fixed base rate $\eta$ by per-config embedding/decoder multipliers and differs across configs only in those two rates. SFT and FFT letters index independent schemes. For each language and strategy we train configurations $\{$A, B, C$\}$ and keep the run with the lowest FLEURS-dev WER, over two seeds (42, 1337) for SFT and three seeds (42, 1337, 2024) for FFT.}
\label{tab:configs}
\end{table*}

\subsection{Hyperparameter Sweep Results}
\label{app:hp_sweep_results}
We conducted a set of Bayesian hyperperameter sweeps for the 17 core languages.
\Cref{tab:hp_sweep} lists the per-language optima for the 17 core languages sorted by FLEURS-test WER (lower is better). 
These optima motivated the discrete \vft configurations used at scale and were not deployed directly. 
We ran a 5-run sweep over a six-dimensional search space: decoder learning rate $\in[10^{-5}, 8\times10^{-5}]$, gradient accumulation $\in\{16,24,32\}$, scheduler $\in\{\text{linear},\text{cosine},\text{exponential}\}$, decay rate $\in\{0.9,0.95,0.99\}$, warmup steps $\in\{100,150\}$, and weight decay $\in[10^{-5},10^{-3}]$. Each run trained for up to 12 epochs with patience 3.

\onecolumn
\begin{table}[h]
\centering
\small
\setlength{\tabcolsep}{3pt}
\begin{tabular}{l r r r l r r r}
\toprule
Language & WER & lr & ga & sched. & decay & warmup & wd \\
\midrule
Italian    &  2.39\% & $1.07{\times}10^{-5}$ & 32 & linear      & 0.90 & 150 & $2.65{\times}10^{-5}$ \\
Spanish    &  3.75\% & $4.41{\times}10^{-5}$ & 32 & cosine      & 0.99 & 150 & $4.93{\times}10^{-5}$ \\
Catalan    &  4.29\% & $1.58{\times}10^{-5}$ & 24 & cosine      & 0.90 & 100 & $1.69{\times}10^{-4}$ \\
Russian    &  5.35\% & $1.77{\times}10^{-5}$ & 24 & linear      & 0.99 & 100 & $6.51{\times}10^{-5}$ \\
Polish     &  5.98\% & $1.87{\times}10^{-5}$ & 24 & linear      & 0.95 & 100 & $4.39{\times}10^{-5}$ \\
Galician   &  6.32\% & $1.65{\times}10^{-5}$ & 16 & linear      & 0.99 & 150 & $6.63{\times}10^{-4}$ \\
German     &  6.47\% & $4.30{\times}10^{-5}$ & 32 & cosine      & 0.95 & 150 & $6.86{\times}10^{-5}$ \\
Ukrainian  &  6.87\% & $1.16{\times}10^{-5}$ & 16 & linear      & 0.90 & 150 & $5.93{\times}10^{-4}$ \\
Cantonese  &  9.39\% & $7.81{\times}10^{-5}$ & 24 & exponential & 0.95 & 150 & $3.36{\times}10^{-4}$ \\
Latvian    & 11.90\% & $2.33{\times}10^{-5}$ & 16 & linear      & 0.99 & 150 & $1.71{\times}10^{-5}$ \\
Arabic     & 13.13\% & $2.03{\times}10^{-5}$ & 32 & cosine      & 0.90 & 100 & $3.40{\times}10^{-5}$ \\
Tajik      & 13.34\% & $4.16{\times}10^{-5}$ & 16 & linear      & 0.95 & 100 & $1.30{\times}10^{-5}$ \\
Swahili    & 24.43\% & $1.57{\times}10^{-5}$ & 32 & linear      & 0.90 & 150 & $5.23{\times}10^{-5}$ \\
Uzbek      & 28.51\% & $6.99{\times}10^{-5}$ & 24 & cosine      & 0.99 & 150 & $1.25{\times}10^{-5}$ \\
Tamil      & 31.22\% & $4.55{\times}10^{-5}$ & 32 & exponential & 0.90 & 100 & $3.87{\times}10^{-5}$ \\
Pashto     & 37.77\% & $7.63{\times}10^{-5}$ & 24 & linear      & 0.99 & 100 & $3.61{\times}10^{-5}$ \\
Georgian   & 51.12\% & $4.48{\times}10^{-5}$ & 24 & cosine      & 0.95 & 100 & $4.64{\times}10^{-4}$ \\
\bottomrule
\end{tabular}
\caption{Per-language optima from a Bayesian hyperparameter pilot over the 17 core languages, sorted by FLEURS-test WER (lower is better). Because per-language search does not scale to 102 languages, these optima were used to \emph{design} the small set of discrete configurations applied in all main experiments (\Cref{tab:configs}); the deployed models do not use per-language hyperparameters. \textbf{lr}: decoder learning rate; \textbf{ga}: gradient accumulation; \textbf{sched.}: LR schedule; \textbf{decay}: scheduler decay rate; \textbf{warmup}: warmup steps; \textbf{wd}: weight decay. Encoder LR is held at $0.3\times$ the decoder LR. All runs use AdamW, micro-batch size~2, and patience-based early stopping on FLEURS-validation.}
\label{tab:hp_sweep}
\end{table}

\twocolumn

\section{Multitask Learning Mixing Ratios}\label{app:mtl_mixing}

\begin{figure}
    \centering
    \includegraphics[width=\linewidth]{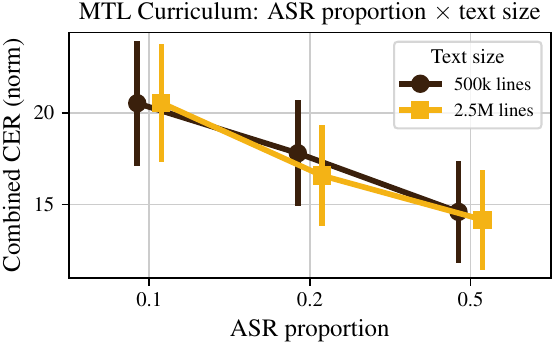}
    \caption{Comparison of \mtl with different proportions of ASR data and text dataset sizes of 500k and 2.5M lines. We use an ASR proportion of 0.5 and a text size of 500k lines in our main experiments.}
    \label{fig:mtl_ratio}
\end{figure}

In our main \fft experiments, we combine \textonly and \asrft batches in a 1:1 ratio, i.e., the proportion of ASR batches is 50\%.
We test the effect of this mixing ratio, conducting additional experiments with ASR proportions of 10\% and 20\%.%
\footnote{We opt for lower percentages of ASR data due to the fact that we generally have less ASR data than \textonly data. %
}
We independently manipulate the length of \mtl to either 500k lines (as in our main experiments) or 2.5M lines. 

Our results, shown in \Cref{fig:mtl_ratio}, indicate that the ASR proportion used in our main experiments is optimal, at least among the settings we evaluate.
More specifically, we find that performance improves with larger ASR proportions.
Furthermore, we find little or no improvement due to increasing the text dataset size from 500k lines (as in our main experiments) to 2.5M lines.
These findings leave open the possibility that higher ASR proportions---or perhaps annealing curricula that increase the ASR proportion towards the end of training---might yield stronger performance.
However, due to the wide range of possible \mtl approaches, we leave a more thorough investigation to future work.

\section{Tokenizer Replacement Strategies}\label{app:tokenizer_strategy}

\begin{figure}
    \centering
    \includegraphics[width=\linewidth]{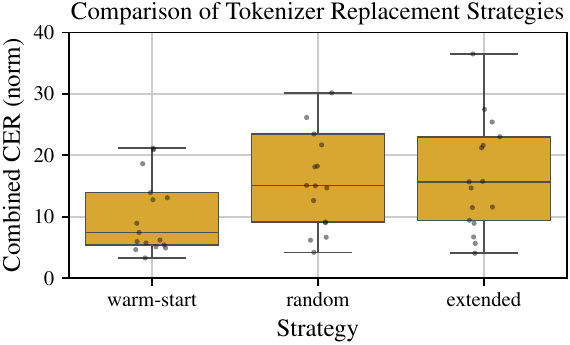}
    \caption{Comparison of tokenizer replacement strategies. We use the warm-start approach in our main experiments. \emph{For presentational reasons, we limit the $y$-axis, excluding some outliers.}}
    \label{fig:tokenizer_strategy}
\end{figure}

We test three strategies for tokenizer replacement. All three strategies begin by training a new BPE tokenizer on a monolingual dataset.
\begin{enumerate}
    \item 
        The \newterm{warm-start} strategy (described in greater detail in \Cref{sec:tk_rep}) entails replacing the original tokenizer entirely with the new tokenizer, and initializing all embeddings (and unembeddings) using the original (un)embeddings or as a sum of the original (un)embeddings making up each new token.
    \item 
        The \newterm{random} strategy replaces the original tokenizer entirely with the new tokenizer, but initializes (un)embeddings randomly.
    \item 
        The \newterm{extended} strategy appends all new tokens to the end of the original tokenizer merge list initializes new embeddings as a sum of the original embeddings.
\end{enumerate}

For each tokenizer replacement strategies, we conduct three training runs for the 17 core languages. 
The results are shown in \Cref{fig:tokenizer_strategy}.
We find that the warm-start strategy adopted in our main experiments is the most effective, yielding a median CER of 7.4.
The remaining random and extended strategies lead to worse performance overall, yielding median CERs of 15.1 and 15.7, respectively.

\section{Out-of-Domain Generalization}\label{app:ood}

\begin{figure*}
    \centering
    \includegraphics[width=\linewidth]{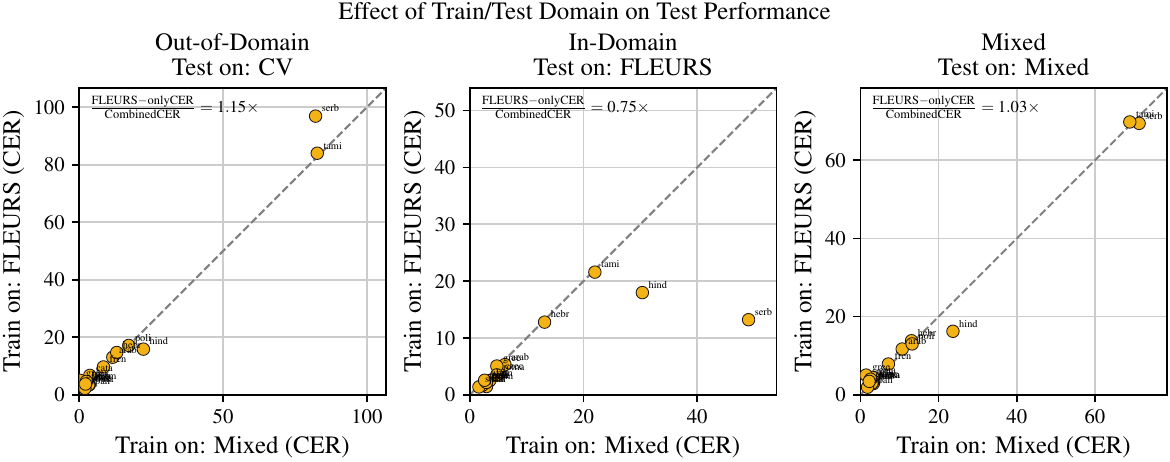}
    \caption{Comparison of in-domain and out-of-domain evaluation. We compare models trained on FLEURS alone to models trained on both Common Voice and Fleurs as in our main experiments. Models are evaluated on Common Voice (out-of-domain), FLEURS (in-domain), and both (mixed).}
    \label{fig:ood}
\end{figure*}

In our main experiments, our train and test sets for most languages are a combination of data from FLEURS and Common Voice.
However, other models we compare against using the same test data may not be trained on these domains, or are trained on a wider set of domains.
Therefore, we conduct out-of-domain evaluations to investigate the extent to which our performance gains are due to the similarity of the train/test domain, rather than language-specific fine-tuning.

We train models on the 17 core languages using \vft on only the FLEURS portion of the training data, and we evaluate on either the Common Voice test data (out-of-domain), the FLEURS test set (in-domain), or both (mixed).
The results in \Cref{fig:ood} reveal a modest in-domain advantage relative to models trained on both FLEURS and Common Voice.
The FLEURS-only model tested OOD is outperformed by a model trained on both domains by a factor of 1.15{$\times$}, while it outperforms the mixed model on the in-domain evaluation by a factor of 1.33{$\times$}. 
On the combined test set, there is little difference, with the mixed models outperforming the FLEURS models by a factor 1.03{$\times$} on average.
We conclude that some of the performance gains over Whisper and external baselines that we observe may be due to better alignment between the fine-tuning data and the test data domains.
However, this alone cannot explain the average performance gain of 2.8{$\times$} that we observe for BuzzASR over Whisper.

\section{Multilingual Fine-Tuning}\label{app:multilingual}

\begin{figure}
    \centering
    \includegraphics[width=\linewidth]{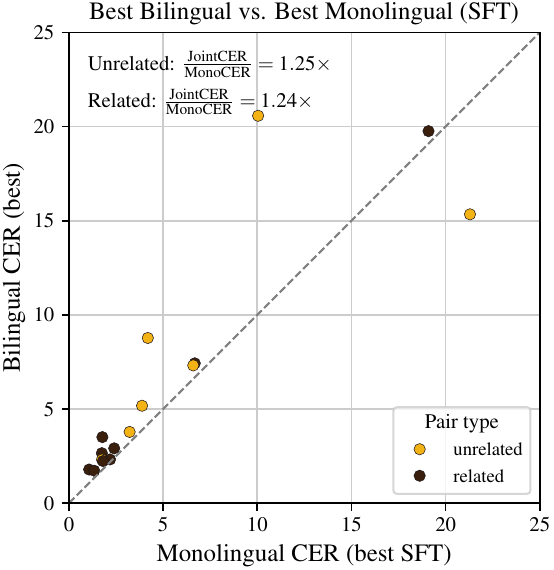}
    \caption{Comparison of monolingual \vft and bilingual \vft for 10 language pairs (5 pairs of related languages and 5 pairs of unrelated languages). \emph{For presentational reasons, we limit the $y$-axis, excluding some outliers.}}
    \label{fig:bilingual}
\end{figure}

The main premise of our work is that monolingual fine-tuning of massively multilingual models can improve language-specific ASR performance.
However, prior work on text-based models showed that limiting multilinguality to a small number of languages can sometimes improve performance over monolingual models \citep{chang-etal-2024-multilinguality}.
We test the initial promise of that approach in our setting by conducting \emph{bilingual} fine-tuning runs.
We select 10 language pairs from our core set of 17 languages.
5 pairs are unrelated (Arabic--Tamil, Cantonese--Georgian, German--Swahili, Latvian--Uzbek, Spanish--Cantonese), and 5 pairs are related (Catalan--Galician, Italian--Spanish, Polish--Russian, Russian--Ukrainian, Tajik--Pashto). 
For each pair, we conduct three \vft training runs with different random seeds.

The results are shown in \Cref{fig:bilingual}.
We find that the best bilingual result is nearly always worse than the best monolingual \vft result for each language in that pair.
Furthermore, we find no advantage to training on related languages as opposed to unrelated languages:
The median improvements of monolingual training over bilingual training are 1.25{$\times$} and 1.24{$\times$} for unrelated and related language pairs, respectively. 
The only exceptions come from the bilingual Cantonese--Georgian models, where we observe a substantial bilingual improvement of 1.38{$\times$} for Cantonese and a modest bilingual improvement of 1.01{$\times$} for Georgian.

\section{Ablation: Multitask Learning Without Tokenizer Replacement}\label{app:mtl_tk_ablation}

\begin{figure}
    \centering
    \includegraphics[width=\linewidth]{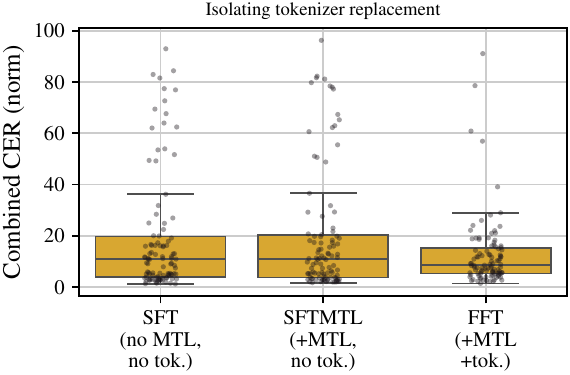}
    \caption{Ablation of tokenizer replacement from the \fft setting. We compare \vft and \fft to runs where we perform \mtl without tokenizer replacement (SFTMTL).}
    \label{fig:mtl_ablation}
\end{figure}

To isolate the effects of \mtl and tokenizer replacement, we conduct additional training runs in which we perform \mtl \emph{without} tokenizer adaptation.%
\footnote{We do not investigate tokenizer adaptation without \mtl because we consider tokenizer adaptation to be likely to fail without the additional \textonly fine-tuning data.}
For each of the 17 core languages, we conduct three additional \mtl runs of this kind.

The results in \Cref{fig:mtl_ablation} compare all \vft and \fft runs against this \mtlonly approach.
In the aggregate, we find that \fft is more effective than either alternative, and there is little difference in performance between \vft and \mtlonly. 

\section{Decoder Latency}
\label{app:latency}

We benchmark per-utterance decoder latency across the three systems
(Whisper zero-shot, \vft, and \fft) to assess whether the tokenizer replacement employed in \fft translates to a wall-clock speedup.
For each (language, system) pair we sample 50 FLEURS-test utterances, and, where available, 50 from Common Voice; warm up with three untimed \texttt{generate()} calls; then time each subsequent call with \texttt{torch.cuda.synchronize} barriers and
\texttt{time.perf\_counter} deltas. 
All three systems use plain greedy decoding (\texttt{num\_beams=1}, no repetition penalty or \textit{no-repeat-ngram} suppression) so the comparison isolates the model and tokenizer rather than decoding heuristics. 
We use FP16, batch size 1, \texttt{max\_new\_tokens=256}, and a single A100-80GB GPU per measurement.

We analyze three quantities per (lang, system): median per-utterance
wall-clock latency in milliseconds, median per-token throughput in
tokens per second, and median number of tokens generated. The first
captures user-facing latency; the second isolates raw decode speed
(which we expect to be constant across systems since all share the
Whisper-large-v3 architecture); the third explains the gap.

\begin{table*}[t]
\centering
\small
\setlength{\tabcolsep}{6pt}
\begin{tabular}{l r r r r r r}
\toprule
& & \multicolumn{3}{c}{\textbf{Latency (ms / utt)}} & \multicolumn{2}{c}{\textbf{Speedup vs.\ Whisper-zs}} \\
\cmidrule(lr){3-5} \cmidrule(lr){6-7}
\textbf{Script} & \textbf{n} & Whisper-zs & SFT & FFT & \textbf{FFT} & SFT \\
\midrule
Latin                     & 61 &  714 &  714 &  448 & \textbf{1.59$\times$} & 1.00$\times$ \\
Cyrillic                  & 10 &  785 &  788 &  449 & \textbf{1.75$\times$} & 1.00$\times$ \\
Arabic                    &  6 &  964 & 1101 &  487 & \textbf{1.98$\times$} & 0.88$\times$ \\
Brahmic                   & 16 & 1611 & 2771 & 1384 & \textbf{1.16$\times$} & 0.58$\times$ \\
Sino-Japanese$^{\dagger}$ &  4 &  502 &  541 &  514 & 0.98$\times$ & 0.93$\times$ \\
Other                     &  5 & 1475 & 1861 &  420 & \textbf{3.51$\times$} & 0.79$\times$ \\
\midrule
\textbf{Global}          & 102 &  785 &  818 &  503 & \textbf{$\sim$1.6$\times$} & 0.96$\times$ \\
\bottomrule
\end{tabular}
\caption{Per-script decoder latency on the combined FLEURS+CV25 test set (idle GPU), median over languages. Speedup is the median per-language ratio of Whisper-zs to system latency. ``Other'' covers unique scripts: Amharic (Ge'ez), Armenian, Georgian, Greek, and Hebrew. $^{\dagger}$Sino-Japanese is roughly latency-neutral; the residual cost is driven by Japanese, which over-generates due to a generation failure.}
\label{tab:latency-by-script}
\end{table*}

Three findings follow. First, per-token throughput is essentially identical across the three systems they share the Whisper-large-v3 decoder (median $\approx$73 tokens/sec) confirming that latency differences arise from sequence length rather than per-token decode speed. Second, FFT's native tokenizer delivers a wall-clock speedup on every script bucket except Sino-Japanese, and each speedup tracks that script's token-count compression: 1.59$\times$ on Latin, 1.75$\times$ on Cyrillic, 1.98$\times$ on Arabic, 1.16$\times$ on Brahmic, and 3.51$\times$ on other languages, for a global median of $\sim$1.6$\times$.

\section{Per-Language Tokenizer Metrics}
\label{app:tokenizer_per_lang}

\begin{figure*}
    \centering
    \includegraphics[width=\linewidth]{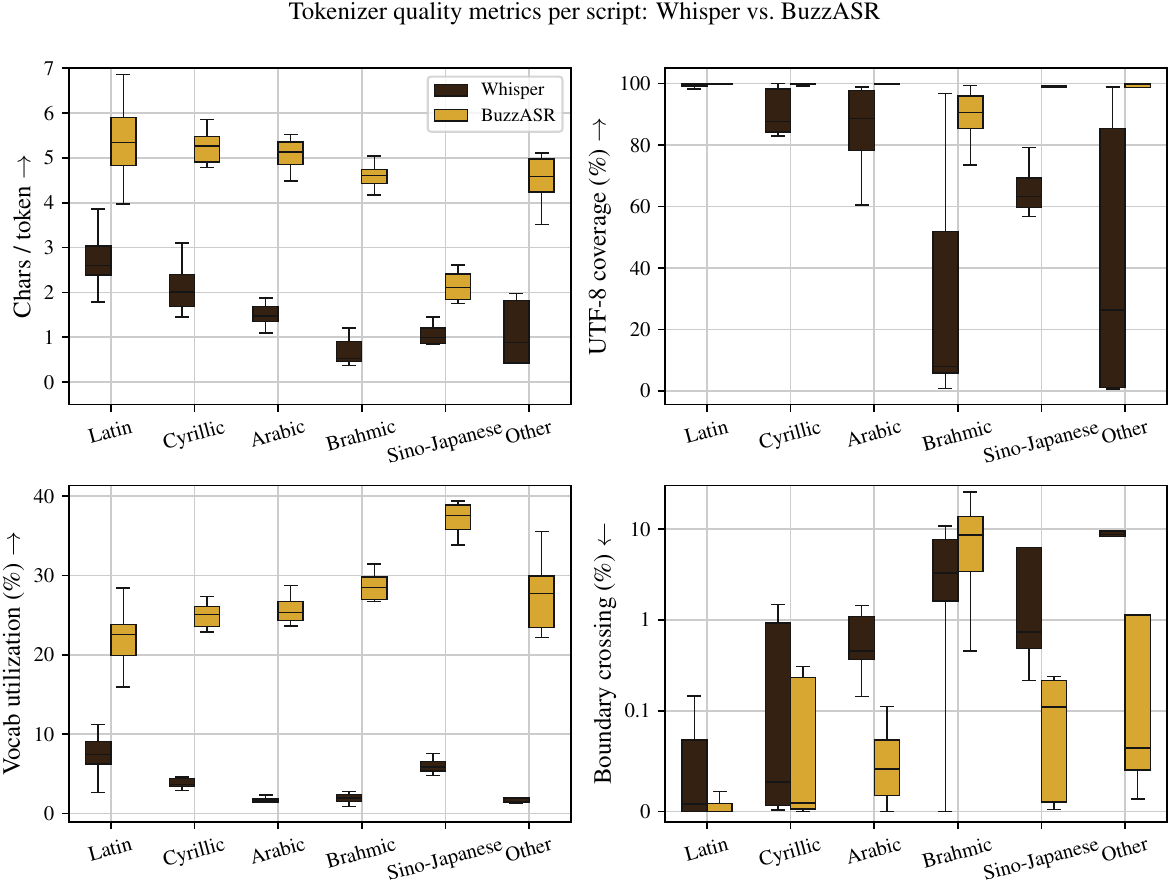}
    \caption{Tokenizer performance metrics for Whisper and BuzzASR, separated by script. Arrows indicate the direction associated with \emph{better} performance.}
    \label{fig:tokenizer_metrics}
\end{figure*}

\Cref{fig:tokenizer_metrics} summarizes per-language tokenizer-quality
metrics for all 102 languages: characters per token, UTF-8 coverage,
vocabulary utilization, and boundary crossing of our per-language BPE
over Whisper's multilingual BPE on a sample of the text corpus used to train the per-language tokenizers.
Detailed results are provided in \Cref{tab:per_lang_tokenizer}.

Nearly across the board, the BuzzASR tokenizers outperform Whisper tokenizers. 
Our tokenizers generally achieve around 5 characters per token, while the Whisper tokenizer achieves less than 2 for all non-Latin/Cyrillic scripts.
The main exception is Sino-Japanese scripts, which are largely logographic. 
In fact, the Whisper tokenizer reaches less than 1 character per token for Brahmic and the rarer scripts (``other'').
Our UTF-8 coverage is typically at 100\%, meaning that when a token sequence is decoded back to text, essentially no characters are corrupted.
Vocab utilization is generally above 20\% for BuzzASR and below 10\% for Whisper. 
For Arabic, Brahmic, and rarer scripts, Whisper vocab utilization is generally below 3\%, meaning 97\% of the vocabulary is unused.
Finally, boundary crossing for BuzzASR is below 0.1\% for the vast majority of languages except those using Brahmic script.
Boundary crossing is the percentage of tokens that contain an internal whitespace character i.e., a single token that spans a word boundary by merging parts of adjacent words
By contrast, for Whisper, boundary crossing only falls below this threshold for languages using Latin and Cyrillic scripts.

\onecolumn
\begin{longtable}{lc ccc ccc c}
\caption{Per-language tokenizer quality metrics. ``ours'' is our per-language BPE, ``W'' is Whisper-large-v3's multilingual BPE. Higher chars/tok and vocab utilization are better; lower tokens/word is better. Kamba's per-language tokenizer is anomalous due to its tiny training corpus and is excluded from aggregate statistics in the main text.}
\label{tab:per_lang_tokenizer}\\
\toprule
& & \multicolumn{2}{c}{\textbf{chars/tok}} & \multicolumn{2}{c}{\textbf{tokens/word}} & \multicolumn{2}{c}{\textbf{vocab util.}} & \textbf{Comp.} \\
\cmidrule(lr){3-4} \cmidrule(lr){5-6} \cmidrule(lr){7-8}
\textbf{Language} & \textbf{Script} & \textbf{ours} & \textbf{W} & \textbf{ours} & \textbf{W} & \textbf{ours} & \textbf{W} & \textbf{gain} \\
\midrule
\endfirsthead
\toprule
& & \multicolumn{2}{c}{\textbf{chars/tok}} & \multicolumn{2}{c}{\textbf{tokens/word}} & \multicolumn{2}{c}{\textbf{vocab util.}} & \textbf{Comp.} \\
\cmidrule(lr){3-4} \cmidrule(lr){5-6} \cmidrule(lr){7-8}
\textbf{Language} & \textbf{Script} & \textbf{ours} & \textbf{W} & \textbf{ours} & \textbf{W} & \textbf{ours} & \textbf{W} & \textbf{gain} \\
\midrule
\endhead
\bottomrule
\endfoot
Afrikaans & Latin & 5.96 & 3.10 & 0.92 & 1.77 & 21.2\% & 7.3\% & 1.93x \\
Amharic & Other & 3.52 & 0.43 & 1.43 & 11.74 & 35.5\% & 1.4\% & 8.23x \\
Arabic & Arabic & 4.78 & 1.87 & 1.22 & 3.10 & 28.7\% & 1.8\% & 2.55x \\
Armenian & Other & 5.11 & 0.88 & 1.43 & 8.28 & 23.4\% & 2.0\% & 5.79x \\
Assamese & Brahmic & 4.52 & 0.51 & 1.40 & 12.42 & 29.2\% & 2.0\% & 8.85x \\
Asturian & Latin & 5.04 & 2.89 & 1.21 & 2.11 & 19.9\% & 10.5\% & 1.74x \\
Azerbaijani & Latin & 5.25 & 2.14 & 1.41 & 3.47 & 24.5\% & 6.2\% & 2.46x \\
Belarusian & Cyrillic & 4.78 & 1.95 & 1.46 & 3.57 & 26.9\% & 3.8\% & 2.45x \\
Bengali & Brahmic & 4.65 & 0.51 & 1.36 & 12.36 & 28.4\% & 1.0\% & 9.11x \\
Bosnian & Latin & 4.82 & 2.53 & 1.40 & 2.66 & 22.6\% & 8.2\% & 1.90x \\
Bulgarian & Cyrillic & 4.88 & 2.47 & 1.28 & 2.52 & 27.3\% & 4.6\% & 1.98x \\
Burmese & Brahmic & 2.54 & 0.37 & 5.12 & 34.75 & 16.9\% & 2.4\% & 6.78x \\
Cantonese & CJK & 1.86 & 0.87 & 28.34 & 60.73 & 33.8\% & 4.8\% & 2.14x \\
Catalan & Latin & 5.79 & 3.45 & 1.01 & 1.69 & 22.6\% & 9.5\% & 1.68x \\
Cebuano & Latin & 5.42 & 2.80 & 1.11 & 2.15 & 18.7\% & 8.9\% & 1.94x \\
Croatian & Latin & 5.35 & 2.39 & 1.28 & 2.86 & 24.4\% & 6.2\% & 2.24x \\
Czech & Latin & 4.72 & 2.30 & 1.32 & 2.71 & 28.4\% & 7.1\% & 2.05x \\
Danish & Latin & 6.07 & 3.05 & 1.00 & 2.00 & 23.4\% & 7.9\% & 1.99x \\
Dutch & Latin & 5.96 & 3.43 & 1.04 & 1.81 & 21.2\% & 8.8\% & 1.74x \\
English & Latin & 5.81 & 4.83 & 0.99 & 1.19 & 21.7\% & 15.5\% & 1.20x \\
Estonian & Latin & 5.61 & 2.57 & 1.34 & 2.93 & 23.5\% & 6.6\% & 2.18x \\
Filipino & Latin & 5.73 & 3.03 & 1.03 & 1.95 & 23.7\% & 8.4\% & 1.89x \\
Finnish & Latin & 7.59 & 2.84 & 1.07 & 2.87 & 22.5\% & 7.0\% & 2.67x \\
French & Latin & 5.80 & 3.77 & 1.04 & 1.60 & 22.9\% & 11.2\% & 1.54x \\
Fulah & Latin & 4.12 & 2.28 & 1.39 & 2.52 & 27.4\% & 10.2\% & 1.81x \\
Galician & Latin & 5.92 & 3.49 & 1.01 & 1.72 & 23.3\% & 9.5\% & 1.69x \\
Georgian & Other & 4.58 & 0.41 & 1.79 & 20.00 & 22.2\% & 1.3\% & 11.17x \\
German & Latin & 5.30 & 3.70 & 1.33 & 1.90 & 20.5\% & 10.9\% & 1.43x \\
Greek & Other & 4.98 & 1.98 & 1.30 & 3.27 & 27.7\% & 3.0\% & 2.52x \\
Gujarati & Brahmic & 4.33 & 0.41 & 1.34 & 14.09 & 30.6\% & 2.4\% & 10.53x \\
Hausa & Latin & 5.59 & 2.66 & 0.98 & 2.06 & 21.2\% & 6.3\% & 2.10x \\
Hebrew & Other & 4.25 & 1.82 & 1.31 & 3.07 & 29.9\% & 2.1\% & 2.33x \\
Hindi & Brahmic & 4.67 & 1.02 & 1.07 & 4.90 & 28.3\% & 2.7\% & 4.59x \\
Hungarian & Latin & 5.35 & 2.43 & 1.36 & 2.98 & 24.9\% & 6.4\% & 2.20x \\
Icelandic & Latin & 5.29 & 2.24 & 1.15 & 2.72 & 25.4\% & 5.8\% & 2.36x \\
Igbo & Latin & 4.86 & 2.00 & 1.03 & 2.51 & 22.4\% & 6.8\% & 2.43x \\
Indonesian & Latin & 5.90 & 3.40 & 1.17 & 2.04 & 23.8\% & 6.9\% & 1.74x \\
Irish & Latin & 6.12 & 2.39 & 0.97 & 2.48 & 22.9\% & 4.9\% & 2.56x \\
Italian & Latin & 5.79 & 3.50 & 1.10 & 1.82 & 23.1\% & 9.2\% & 1.65x \\
Japanese & CJK & 2.34 & 1.12 & 14.58 & 30.44 & 38.7\% & 6.2\% & 2.09x \\
Javanese & Latin & 4.56 & 2.85 & 1.37 & 2.20 & 18.6\% & 6.9\% & 1.60x \\
Kabuverdianu & Latin & 4.64 & 2.65 & 1.12 & 1.96 & 25.5\% & 10.0\% & 1.75x \\
Kamba & Latin & 47.45 & 2.18 & 0.13 & 2.78 & 4.1\% & 6.3\% & 21.72x \\
Kannada & Brahmic & 4.18 & 0.61 & 1.91 & 13.12 & 28.5\% & 1.4\% & 6.86x \\
Kazakh & Cyrillic & 5.74 & 1.58 & 1.34 & 4.89 & 23.5\% & 3.5\% & 3.64x \\
Khmer & Brahmic & 4.70 & 0.48 & 4.03 & 39.80 & 22.0\% & 1.5\% & 9.87x \\
Korean & CJK & 2.61 & 1.45 & 1.64 & 2.95 & 36.5\% & 7.6\% & 1.80x \\
Kyrgyz & Cyrillic & 5.86 & 1.67 & 1.32 & 4.61 & 23.5\% & 3.4\% & 3.50x \\
Lao & Brahmic & 5.04 & 0.37 & 3.09 & 42.50 & 26.8\% & 2.8\% & 13.77x \\
Latvian & Latin & 4.86 & 2.20 & 1.46 & 3.23 & 25.1\% & 7.7\% & 2.21x \\
Lingala & Latin & 4.30 & 2.15 & 1.40 & 2.79 & 16.4\% & 8.3\% & 2.00x \\
Lithuanian & Latin & 5.42 & 2.39 & 1.38 & 3.13 & 23.6\% & 6.9\% & 2.27x \\
Luganda & Latin & 5.98 & 2.35 & 1.37 & 3.47 & 15.9\% & 2.7\% & 2.54x \\
Luo & Latin & 4.01 & 2.62 & 1.31 & 2.01 & 24.9\% & 9.7\% & 1.53x \\
Luxembourgish & Latin & 4.88 & 2.86 & 1.31 & 2.24 & 18.8\% & 9.6\% & 1.71x \\
Macedonian & Cyrillic & 5.00 & 2.27 & 1.24 & 2.72 & 26.3\% & 4.1\% & 2.20x \\
Malay & Latin & 6.86 & 3.43 & 0.96 & 1.91 & 19.0\% & 5.5\% & 2.00x \\
Malayalam & Brahmic & 4.57 & 0.48 & 2.04 & 19.53 & 26.7\% & 1.8\% & 9.58x \\
Maltese & Latin & 5.97 & 2.08 & 1.30 & 3.73 & 21.5\% & 5.7\% & 2.87x \\
Mandarin & CJK & 1.76 & 0.85 & 28.00 & 57.88 & 39.4\% & 5.5\% & 2.07x \\
Maori & Latin & 4.50 & 2.37 & 1.15 & 2.19 & 11.9\% & 5.4\% & 1.90x \\
Marathi & Brahmic & 4.73 & 0.90 & 1.39 & 7.31 & 29.3\% & 1.4\% & 5.26x \\
Mongolian & Cyrillic & 5.49 & 1.45 & 1.16 & 4.41 & 22.9\% & 3.1\% & 3.80x \\
Nepali & Brahmic & 4.58 & 0.91 & 1.36 & 6.88 & 31.5\% & 1.9\% & 5.05x \\
Norwegian & Latin & 6.03 & 3.04 & 1.00 & 1.99 & 23.8\% & 8.2\% & 1.99x \\
Nyanja & Latin & 4.47 & 2.62 & 1.53 & 2.61 & 18.2\% & 9.1\% & 1.71x \\
Occitan & Latin & 4.92 & 2.78 & 1.22 & 2.15 & 20.9\% & 9.6\% & 1.77x \\
Oriya & Brahmic & 4.84 & 0.38 & 1.33 & 16.93 & 29.7\% & 0.9\% & 12.73x \\
Oromo & Latin & 4.50 & 2.47 & 1.60 & 2.92 & 25.5\% & 8.4\% & 1.82x \\
Pashto & Arabic & 5.10 & 1.34 & 0.90 & 3.42 & 27.1\% & 0.9\% & 3.81x \\
Persian & Arabic & 5.43 & 1.57 & 0.91 & 3.17 & 23.6\% & 1.7\% & 3.47x \\
Polish & Latin & 5.02 & 3.04 & 1.39 & 2.29 & 26.8\% & 8.5\% & 1.65x \\
Portuguese & Latin & 5.79 & 3.80 & 1.03 & 1.57 & 22.8\% & 10.1\% & 1.52x \\
Punjabi & Brahmic & 4.79 & 0.54 & 1.07 & 9.55 & 28.2\% & 2.1\% & 8.95x \\
Romanian & Latin & 5.29 & 2.81 & 1.15 & 2.17 & 24.8\% & 8.5\% & 1.88x \\
Russian & Cyrillic & 5.34 & 3.10 & 1.35 & 2.32 & 25.3\% & 6.4\% & 1.72x \\
Sepedi & Latin & 5.30 & 2.56 & 1.12 & 2.31 & 12.1\% & 5.8\% & 2.07x \\
Serbian & Cyrillic & 4.87 & 2.06 & 1.31 & 3.09 & 24.9\% & 3.4\% & 2.36x \\
Shona & Latin & 4.53 & 2.71 & 1.62 & 2.71 & 16.7\% & 8.2\% & 1.67x \\
Sindhi & Arabic & 5.16 & 1.38 & 0.97 & 3.64 & 24.1\% & 2.3\% & 3.75x \\
Slovak & Latin & 5.04 & 2.36 & 1.28 & 2.73 & 26.2\% & 7.0\% & 2.14x \\
Slovenian & Latin & 5.57 & 2.56 & 1.14 & 2.48 & 24.6\% & 6.3\% & 2.18x \\
Somali & Latin & 6.43 & 2.48 & 1.00 & 2.59 & 22.0\% & 4.8\% & 2.60x \\
Sorani-kurdish & Arabic & 4.49 & 1.09 & 1.44 & 5.92 & 25.0\% & 1.5\% & 4.10x \\
Spanish & Latin & 5.97 & 3.86 & 1.00 & 1.55 & 22.4\% & 10.7\% & 1.55x \\
Swahili & Latin & 5.96 & 2.56 & 1.05 & 2.46 & 22.2\% & 6.1\% & 2.33x \\
Swedish & Latin & 5.58 & 3.23 & 1.10 & 1.90 & 23.8\% & 8.6\% & 1.73x \\
Tajik & Cyrillic & 5.44 & 1.73 & 1.22 & 3.83 & 23.9\% & 2.9\% & 3.14x \\
Tamil & Brahmic & 4.99 & 1.16 & 1.73 & 7.46 & 27.0\% & 1.7\% & 4.32x \\
Telugu & Brahmic & 4.34 & 0.56 & 1.75 & 13.66 & 30.1\% & 2.1\% & 7.81x \\
Thai & Brahmic & 4.46 & 1.21 & 4.20 & 15.52 & 30.0\% & 2.6\% & 3.69x \\
Turkish & Latin & 5.43 & 2.99 & 1.37 & 2.49 & 23.8\% & 7.4\% & 1.81x \\
Ukrainian & Cyrillic & 5.20 & 2.44 & 1.35 & 2.87 & 25.6\% & 4.6\% & 2.13x \\
Umbundu & Latin & 4.58 & 2.56 & 1.27 & 2.27 & 23.7\% & 8.2\% & 1.79x \\
Urdu & Arabic & 5.52 & 1.73 & 0.83 & 2.66 & 25.6\% & 1.4\% & 3.20x \\
Uzbek & Latin & 5.31 & 2.48 & 1.38 & 2.96 & 18.3\% & 4.6\% & 2.15x \\
Vietnamese & Latin & 5.26 & 2.60 & 0.85 & 1.72 & 24.4\% & 4.1\% & 2.02x \\
Welsh & Latin & 5.90 & 2.43 & 0.97 & 2.36 & 21.5\% & 5.8\% & 2.43x \\
Wolof & Latin & 4.78 & 2.43 & 1.04 & 2.04 & 16.8\% & 6.3\% & 1.96x \\
Xhosa & Latin & 4.65 & 2.48 & 1.84 & 3.44 & 18.7\% & 6.0\% & 1.87x \\
Yoruba & Latin & 4.83 & 1.78 & 0.98 & 2.67 & 20.8\% & 5.2\% & 2.72x \\
Zulu & Latin & 3.98 & 2.57 & 2.09 & 3.25 & 18.5\% & 9.5\% & 1.55x \\
\end{longtable}

\twocolumn

\section{Full Evaluation Results} \label{app:full_results}

Table \ref{tab:test_all_languages_cer} and \ref{tab:test_all_languages_wer} provide the full CER and WER results for all languages on the combined FLEURS\,+\,Common Voice test set (FLEURS only for the 25 languages not covered by Common Voice).

\onecolumn
\small
\setlength{\tabcolsep}{4pt}
\begin{longtable}{p{2.8cm} ccc cccccc}
\caption{Test-set CER (\%) on the test for all languages, which consists of the combined CV and FLEURS datasets when available and just FLEURS elsewhere. Lower is better. \textbf{Bold} indicates the best score within the Whisper group; \underline{underline} indicates the overall best score. Each method uses the configuration selected on FLEURS validation WER. Scores are normalised CER, except Cohere and Qwen3, which are raw. $^\dagger$\,No Common Voice evaluation was available for this system and language, so the score is on FLEURS only.}
\label{tab:test_all_languages_cer} \\
\toprule
& \multicolumn{3}{c}{\textbf{Ours}} & \multicolumn{6}{c}{\textbf{Baselines}} \\
\cmidrule(lr){2-4} \cmidrule(lr){5-10}
\textbf{Language}
  & {\textbf{FFT}} & {\textbf{SFT}} & {\textbf{SFTMTL}}& {\textbf{Whisper}}
  & {\textbf{Omni 1B}} & {\textbf{Omni 7B}} & {\textbf{MMS}} & {\textbf{Cohere}} & {\textbf{Qwen3}} \\
\midrule
\endfirsthead
\multicolumn{10}{c}{\tablename\ \thetable\ -- continued} \\
\toprule
& \multicolumn{3}{c}{\textbf{Ours}} & \multicolumn{6}{c}{\textbf{Baselines}} \\
\cmidrule(lr){2-4} \cmidrule(lr){5-10}
\textbf{Language}
  & {\textbf{FFT}} & {\textbf{SFT}} & {\textbf{SFTMTL}}& {\textbf{Whisper}}
  & {\textbf{Omni 1B}} & {\textbf{Omni 7B}} & {\textbf{MMS}} & {\textbf{Cohere}} & {\textbf{Qwen3}} \\
\midrule
\endhead
\midrule
\multicolumn{10}{r}{\textit{Continued on next page}} \\
\endfoot
\bottomrule
\endlastfoot
  Afrikaans & 6.72 & \textbf{5.35} & 7.63 & 10.88 & 6.69 & \underline{5.24} & 6.74 & 29.84 & 29.82 \\
\addlinespace[2pt]
  Amharic & \textbf{9.07} & 49.39 & 51.04 & 191.35 & 12.15 & \underline{7.96} & 10.51 & 141.39 & 120.99 \\
\addlinespace[2pt]
  Arabic & 12.52 & 11.74 & \underline{\textbf{10.76}} & 11.82 & 13.83 & 11.60 & 11.31 & 11.99 & 11.27 \\
\addlinespace[2pt]
  Armenian & \underline{\textbf{2.28}} & 13.50 & 17.38 & 15.53 & 5.11 & 3.92 & 5.08 & 118.05 & 93.46 \\
\addlinespace[2pt]
  Assamese & \textbf{11.83} & 62.43 & 62.94 & 80.72 & 10.05 & \underline{7.79} & 9.50 & 121.67 & 96.03 \\
\addlinespace[2pt]
  Asturian & 5.90 & \underline{\textbf{4.89}} & 5.23 & 14.65 & 7.52 & 5.89 & 5.34 & 18.76 & 18.42 \\
\addlinespace[2pt]
  Azerbaijani & 8.34 & \textbf{5.22} & 5.70 & 5.68 & 5.82 & \underline{4.66} & 5.39$^\dagger$ & 92.58 & 34.30 \\
\addlinespace[2pt]
  Belarusian & 19.18 & \textbf{2.75} & 3.29 & 10.95 & 3.52 & \underline{2.14} & 3.37 & 99.79 & 36.56 \\
\addlinespace[2pt]
  Bengali & \textbf{7.79} & 53.91 & 67.33 & 34.04 & 6.58 & \underline{4.01} & 8.40 & 109.42 & 93.88 \\
\addlinespace[2pt]
  Bosnian & 6.37 & \textbf{3.59} & 3.78 & 3.87 & 4.91 & \underline{3.07} & 3.51 & 58.18 & 42.67 \\
\addlinespace[2pt]
  Bulgarian & 2.60 & \underline{\textbf{1.21}} & 1.68 & 4.19 & 6.66 & 4.42 & 3.78 & 98.44 & 30.56 \\
\addlinespace[2pt]
  Burmese & \textbf{24.01} & 84.38 & 82.26 & 123.32 & 11.14 & \underline{8.27} & 19.65 & 105.19 & 98.34 \\
\addlinespace[2pt]
  Cantonese & 19.07 & 13.01 & \underline{\textbf{12.77}} & 36.44 & 33.08 & 27.73 & 60.88$^\dagger$ & 69.27 & 43.15 \\
\addlinespace[2pt]
  Catalan & 4.21 & 4.50 & \textbf{4.20} & 4.32 & 3.76 & \underline{2.64} & 3.49 & 30.77 & 29.28 \\
\addlinespace[2pt]
  Cebuano & 6.24 & 5.91 & \textbf{5.85} & 15.42 & 5.41 & 4.48 & \underline{4.33} & 61.72 & 16.67 \\
\addlinespace[2pt]
  Croatian & 6.05 & \underline{\textbf{3.09}} & 3.61 & 5.11 & 24.36 & 23.34 & 3.38 & 56.73 & 72.08 \\
\addlinespace[2pt]
  Czech & 5.17 & 3.41 & \textbf{3.00} & 3.07 & 3.50 & \underline{2.65} & 2.88 & 51.85 & 9.65 \\
\addlinespace[2pt]
  Danish & 14.90 & 16.21 & 14.55 & \underline{\textbf{4.74}} & 6.41 & 4.74 & 6.60 & 58.33 & 11.36 \\
\addlinespace[2pt]
  Dutch & 3.69 & \underline{\textbf{1.49}} & 1.97 & 1.98 & 3.48 & 2.91 & 3.29 & 3.04 & 3.63 \\
\addlinespace[2pt]
  English & 4.39 & 4.84 & 4.60 & \underline{\textbf{4.34}} & 7.20 & 5.55 & 6.60 & 6.99 & 5.95 \\
\addlinespace[2pt]
  Estonian & \underline{\textbf{1.40}} & 1.96 & 3.51 & 6.19 & 4.46 & 3.30 & 3.73 & 69.25 & 37.42 \\
\addlinespace[2pt]
  Filipino & 5.63 & 5.04 & 5.07 & \textbf{4.46} & 4.05 & \underline{3.32} & 3.68 & 66.84 & 11.23 \\
\addlinespace[2pt]
  Finnish & 28.96 & 31.79 & 31.75 & \underline{\textbf{1.89}} & 2.96 & 2.26 & 2.27 & 66.30 & 8.36 \\
\addlinespace[2pt]
  French & \textbf{6.68} & 7.89 & 7.29 & 7.65 & 7.90 & 6.22 & 7.00 & \underline{5.79} & 6.65 \\
\addlinespace[2pt]
  Fulah & \textbf{16.04} & 16.60 & 17.17 & 33.12 & 15.12 & 15.05 & \underline{14.12} & 81.07 & 60.56 \\
\addlinespace[2pt]
  Galician & 2.48 & \underline{\textbf{1.48}} & 2.91 & 3.84 & 2.96 & 2.01 & 2.73 & 17.13 & 17.86 \\
\addlinespace[2pt]
  Georgian & \textbf{12.54} & 72.65 & 77.96 & 19.43 & 3.73 & \underline{2.39} & 4.93 & 114.39 & 94.31 \\
\addlinespace[2pt]
  German & 5.55 & 3.20 & 2.87 & \textbf{2.77} & 3.86 & \underline{2.75} & 3.78 & 5.40 & 5.88 \\
\addlinespace[2pt]
  Greek & 9.24 & 5.05 & \underline{\textbf{3.21}} & 4.37 & 8.84 & 6.57 & 5.52 & 6.42 & 16.29 \\
\addlinespace[2pt]
  Gujarati & \textbf{8.39} & 77.40 & 79.74 & 20.17 & 5.87 & \underline{4.86} & 6.42 & 113.92 & 92.34 \\
\addlinespace[2pt]
  Hausa & 7.27 & \underline{\textbf{5.50}} & 7.31 & 33.69 & 10.05 & 8.33 & 6.61 & 71.98 & 58.88 \\
\addlinespace[2pt]
  Hebrew & 10.91 & \underline{\textbf{10.89}} & 12.13 & 12.73 & 18.48 & 16.13 & 18.62 & 112.99 & 44.75 \\
\addlinespace[2pt]
  Hindi & 14.32 & 16.23 & 14.76 & \textbf{11.92} & 38.77 & 46.00 & 8.90 & 102.72 & \underline{7.28} \\
\addlinespace[2pt]
  Hungarian & 2.71 & \textbf{2.68} & 3.76 & 3.59 & 3.87 & \underline{2.62} & 3.90 & 81.08 & 13.52 \\
\addlinespace[2pt]
  Icelandic & \textbf{7.45} & 9.00 & 9.74 & 11.25 & 6.00 & \underline{3.60} & 8.29 & 70.96 & 51.12 \\
\addlinespace[2pt]
  Igbo & \textbf{13.97} & 17.26 & 17.30 & 46.49 & 18.07 & 16.05 & \underline{13.83} & 82.37 & 53.56 \\
\addlinespace[2pt]
  Indonesian & 2.67 & 2.63 & \underline{\textbf{1.99}} & 2.36 & 4.61 & 3.84 & 3.54 & 67.96 & 4.28 \\
\addlinespace[2pt]
  Irish & 16.06 & \underline{\textbf{15.70}} & 18.28 & 77.11 & 26.71 & 20.83 & 25.05 & 79.66 & 71.60 \\
\addlinespace[2pt]
  Italian & 5.24 & 2.86 & \textbf{2.53} & 2.74 & 3.46 & \underline{2.44} & 2.71 & 3.60 & 4.06 \\
\addlinespace[2pt]
  Japanese & \underline{\textbf{21.08}} & 24.84 & 23.00 & 23.34 & 32.71 & 28.24 & 36.22 & 27.08 & 23.17 \\
\addlinespace[2pt]
  Javanese & 5.78 & \textbf{5.15} & 5.19 & 25.13 & 5.08 & \underline{4.07} & 4.94 & 84.71 & 25.31 \\
\addlinespace[2pt]
  Kabuverdianu & 5.57 & \textbf{5.03} & 5.53 & 33.02 & 5.52 & \underline{4.17} & 4.26 & 37.56 & 37.78 \\
\addlinespace[2pt]
  Kamba & 18.79 & \textbf{15.92} & 18.09 & 41.55 & 13.21 & \underline{10.97} & 11.73 & 88.27 & 48.72 \\
\addlinespace[2pt]
  Kannada & \textbf{10.29} & 51.67 & 50.59 & 18.72 & 5.10 & \underline{4.07} & 5.25 & 112.65 & 101.46 \\
\addlinespace[2pt]
  Kazakh & \textbf{7.47} & 9.08 & 8.94 & 9.75 & 4.14 & \underline{3.74} & 3.78 & 110.69 & 85.55 \\
\addlinespace[2pt]
  Khmer & \textbf{25.99} & 82.93 & 81.65 & 103.19 & 13.29 & \underline{9.87} & 18.26 & 134.86 & 103.11 \\
\addlinespace[2pt]
  Korean & 6.89 & \underline{\textbf{4.61}} & 5.16 & 5.72 & 13.41 & 11.19 & 20.30 & 9.34 & 7.39 \\
\addlinespace[2pt]
  Kyrgyz & 22.13 & \textbf{11.16} & 13.44 & 34.68 & 6.37 & 4.69 & \underline{4.65} & 105.78 & 89.50 \\
\addlinespace[2pt]
  Lao & \textbf{20.95} & 81.55 & 81.15 & 61.09 & 21.79 & \underline{16.45} & 22.31 & 128.11 & 100.29 \\
\addlinespace[2pt]
  Latvian & 5.45 & \textbf{3.87} & 5.24 & 5.90 & 4.93 & \underline{3.63} & 3.78 & 69.80 & 58.37 \\
\addlinespace[2pt]
  Lingala & 9.13 & \textbf{7.42} & 8.39 & 18.58 & 5.47 & 4.49 & \underline{4.43} & 62.74 & 30.39 \\
\addlinespace[2pt]
  Lithuanian & 2.42 & \underline{\textbf{2.11}} & 2.73 & 6.69 & 6.46 & 4.75 & 3.81 & 65.91 & 79.42 \\
\addlinespace[2pt]
  Luganda & 28.07 & \textbf{24.97} & 29.27 & 43.65 & 8.64 & \underline{6.87} & 7.25 & 63.66 & 46.61 \\
\addlinespace[2pt]
  Luo & 91.07 & 76.92 & 77.23 & \textbf{34.34} & 8.49 & \underline{7.32} & 8.66 & 45.40 & 48.84 \\
\addlinespace[2pt]
  Luxembourgish & \textbf{8.55} & 10.78 & 11.16 & 29.88 & 10.62 & \underline{6.95} & 8.75 & 41.34 & 41.73 \\
\addlinespace[2pt]
  Macedonian & 2.89 & \underline{\textbf{1.31}} & 1.96 & 5.82 & 4.58 & 3.49 & 2.63 & 98.09 & 17.14 \\
\addlinespace[2pt]
  Malay & 3.94 & 2.64 & 2.54 & \underline{\textbf{2.32}} & 4.21 & 3.18 & 3.62$^\dagger$ & 82.04 & 6.72 \\
\addlinespace[2pt]
  Malayalam & \textbf{15.38} & 63.98 & 62.22 & 87.87 & 12.81 & 9.36 & \underline{6.54} & 105.02 & 101.27 \\
\addlinespace[2pt]
  Maltese & 3.41 & \underline{\textbf{2.66}} & 3.60 & 26.35 & 6.55 & 4.66 & 3.75 & 69.83 & 68.89 \\
\addlinespace[2pt]
  Mandarin & 19.17 & 22.44 & 21.63 & \underline{\textbf{15.31}} & 32.40 & 26.80 & 60.94$^\dagger$ & 40.38 & 52.29$^\dagger$ \\
\addlinespace[2pt]
  Maori & \textbf{10.06} & 10.90 & 11.34 & 17.20 & 9.74 & \underline{7.82} & 9.03 & 74.62 & 42.22 \\
\addlinespace[2pt]
  Marathi & 56.91 & 49.19 & 55.48 & \textbf{20.26} & 7.87 & \underline{6.21} & 7.59 & 106.26 & 29.30 \\
\addlinespace[2pt]
  Mongolian & \underline{\textbf{5.21}} & 9.42 & 11.09 & 38.23 & 12.70 & 8.48 & 9.71 & 102.00 & 90.02 \\
\addlinespace[2pt]
  Nepali & \textbf{12.23} & 16.23 & 20.72 & 20.92 & 12.12 & 11.95 & \underline{8.78}$^\dagger$ & 114.36 & 40.12 \\
\addlinespace[2pt]
  Northern Sotho & \textbf{11.79} & 12.91 & 13.15 & 55.10 & 9.48 & \underline{6.38} & 7.35 & 82.26 & 59.92 \\
\addlinespace[2pt]
  Norwegian & 3.38 & 2.78 & \underline{\textbf{2.09}} & 6.37 & 6.00 & 5.15 & 6.57 & 54.22 & 23.16 \\
\addlinespace[2pt]
  Nyanja & \textbf{10.99} & 11.19 & 11.28 & 39.62 & 9.31 & \underline{7.03} & 7.30 & 95.30 & 47.13 \\
\addlinespace[2pt]
  Occitan & \underline{\textbf{10.24}} & 11.85 & 10.73 & 24.07 & 13.34 & 10.46 & 10.64 & 38.44 & 36.27 \\
\addlinespace[2pt]
  Oriya & \textbf{16.33} & 92.95 & 96.24 & 77.97 & 15.25 & 14.22 & \underline{9.01} & 114.41 & 97.05 \\
\addlinespace[2pt]
  Oromo & \textbf{18.59} & 20.01 & 19.77 & 45.25 & 18.78 & \underline{16.52} & 17.41 & 81.66 & 75.81 \\
\addlinespace[2pt]
  Pashto & 17.41 & \textbf{11.10} & 15.54 & 36.45 & 11.29 & \underline{9.86} & 20.10 & 88.61 & 63.01 \\
\addlinespace[2pt]
  Persian & \underline{\textbf{4.71}} & 36.19 & 36.56 & 11.98 & 9.22 & 7.04 & 6.85 & 99.00 & 17.11 \\
\addlinespace[2pt]
  Polish & 7.75 & 13.02 & 12.89 & \underline{\textbf{2.84}} & 4.39 & 3.72 & 3.05 & 5.44 & 8.03 \\
\addlinespace[2pt]
  Portuguese & 3.85 & 3.65 & \underline{\textbf{2.80}} & 3.22 & 4.84 & 3.94 & 3.91 & 5.17 & 5.68 \\
\addlinespace[2pt]
  Punjabi & \underline{\textbf{8.84}} & 53.50 & 48.77 & 42.33 & 19.31 & 23.90 & 9.62 & 113.03 & 87.58 \\
\addlinespace[2pt]
  Romanian & 5.44 & 3.12 & \underline{\textbf{2.90}} & 3.10 & 4.33 & 3.60 & 3.14 & 48.48 & 13.27 \\
\addlinespace[2pt]
  Russian & 4.56 & 2.97 & \underline{\textbf{2.78}} & 2.78 & 5.62 & 3.83 & 5.46 & 99.01 & 5.37 \\
\addlinespace[2pt]
  Serbian & 60.85 & 69.42 & 65.25 & \textbf{58.61} & 35.85 & \underline{32.29} & 89.44$^\dagger$ & 85.50 & 73.78 \\
\addlinespace[2pt]
  Shona & 8.79 & \textbf{8.42} & 8.79 & 26.03 & 4.65 & \underline{3.72} & 4.42 & 88.55 & 40.67 \\
\addlinespace[2pt]
  Sindhi & \textbf{11.59} & 16.10 & 15.94 & 84.65 & 9.72 & \underline{7.74} & 8.53 & 122.37 & 92.38 \\
\addlinespace[2pt]
  Slovak & \underline{\textbf{2.19}} & 4.60 & 2.75 & 5.22 & 7.18 & 6.09 & 2.96 & 44.93 & 25.23 \\
\addlinespace[2pt]
  Slovenian & 4.20 & 2.31 & \underline{\textbf{2.14}} & 7.13 & 9.85 & 9.07 & 4.67 & 56.26 & 73.65 \\
\addlinespace[2pt]
  Somali & \textbf{18.24} & 19.12 & 19.05 & 34.49 & \underline{12.12} & 12.18 & 13.41 & 92.37 & 88.48 \\
\addlinespace[2pt]
  Sorani Kurdish & \textbf{7.43} & 12.71 & 19.85 & 55.24 & 9.32 & \underline{7.02} & 8.41 & 77.66 & 49.32 \\
\addlinespace[2pt]
  Spanish & 3.82 & 1.95 & 2.04 & \underline{\textbf{1.77}} & 2.72 & 2.05 & 2.54 & 3.65 & 3.77 \\
\addlinespace[2pt]
  Swahili & \textbf{12.69} & 17.87 & 19.62 & 22.90 & 7.57 & \underline{6.69} & 7.01 & 55.39 & 32.84 \\
\addlinespace[2pt]
  Swedish & 2.20 & 3.22 & \underline{\textbf{2.14}} & 2.79 & 5.85 & 4.24 & 5.54 & 56.94 & 9.36 \\
\addlinespace[2pt]
  Tajik & 7.35 & \textbf{5.61} & 6.31 & 28.34 & 6.06 & 6.64 & \underline{4.74} & 100.40 & 87.43 \\
\addlinespace[2pt]
  Tamil & 78.57 & 67.64 & 78.42 & \textbf{12.28} & 9.58 & \underline{7.45} & 8.34 & 98.02 & 97.70 \\
\addlinespace[2pt]
  Telugu & \textbf{13.03} & 62.04 & 60.61 & 74.67 & 9.62 & \underline{8.64} & 9.33 & 110.49 & 100.95 \\
\addlinespace[2pt]
  Thai & 39.08 & 28.38 & 27.58 & \textbf{7.52} & 8.87 & \underline{6.63} & 12.11 & 131.03 & 7.11 \\
\addlinespace[2pt]
  Turkish & 4.72 & 4.00 & \underline{\textbf{3.76}} & 4.21 & 8.71 & 6.03 & 4.78 & 81.29 & 6.07 \\
\addlinespace[2pt]
  Ukrainian & 8.12 & 3.44 & \underline{\textbf{2.57}} & 3.21 & 5.52 & 3.69 & 4.42 & 98.75 & 45.06 \\
\addlinespace[2pt]
  Umbundu & 22.00 & \textbf{19.91} & 20.42 & 48.55 & 12.06 & \underline{11.10} & 13.64 & 69.76 & 48.94 \\
\addlinespace[2pt]
  Urdu & 14.49 & 9.24 & 9.46 & \underline{\textbf{8.08}} & 60.60 & 83.07 & 9.96 & 94.42 & 86.79 \\
\addlinespace[2pt]
  Uzbek & 5.94 & \underline{\textbf{3.86}} & 6.63 & 28.45 & 11.64 & 8.71 & 6.00$^\dagger$ & 80.52 & 87.41 \\
\addlinespace[2pt]
  Vietnamese & 6.39 & 5.38 & 5.43 & \underline{\textbf{3.90}} & 7.28 & 5.54 & 12.07 & 8.19 & 6.71 \\
\addlinespace[2pt]
  Welsh & \underline{\textbf{5.24}} & 5.27 & 5.67 & 12.88 & 11.70 & 7.41 & 9.57 & 72.00 & 66.31 \\
\addlinespace[2pt]
  Wolof & \textbf{15.67} & 16.05 & 16.80 & 79.82 & 12.34 & \underline{11.40} & 11.70 & 90.57 & 62.91 \\
\addlinespace[2pt]
  Xhosa & 11.69 & \textbf{11.48} & 11.53 & 39.61 & 8.40 & \underline{6.62} & 6.87 & 102.44 & 50.66 \\
\addlinespace[2pt]
  Yoruba & \textbf{23.67} & 26.92 & 29.20 & 47.96 & 17.36 & \underline{16.71} & 17.91 & 94.82 & 62.62 \\
\addlinespace[2pt]
  Zulu & 11.95 & \textbf{11.86} & 12.15 & 41.29 & 6.89 & \underline{5.58} & 6.42 & 85.32 & 45.85 \\
\end{longtable}
\twocolumn
 
\clearpage
\onecolumn
\small
\setlength{\tabcolsep}{4pt}
\begin{longtable}{p{2.8cm} ccc cccccc}
\caption{Test-set WER (\%) on the test for all languages, which consists of the combined CV and FLEURS datasets when available and just FLEURS elsewhere. Lower is better. \textbf{Bold} indicates the best score within the Whisper group; \underline{underline} indicates the overall best score. Each method uses the configuration selected on FLEURS validation WER. Scores are normalised WER, except Cohere and Qwen3, which are raw. $^\dagger$\,No Common Voice evaluation was available for this system and language, so the score is on FLEURS only.}
\label{tab:test_all_languages_wer} \\
\toprule
& \multicolumn{3}{c}{\textbf{Ours}} & \multicolumn{6}{c}{\textbf{Baselines}} \\
\cmidrule(lr){2-4} \cmidrule(lr){5-10}
\textbf{Language}
  & {\textbf{FFT}} & {\textbf{SFT}} & {\textbf{SFTMTL}}& {\textbf{Whisper}}
  & {\textbf{Omni 1B}} & {\textbf{Omni 7B}} & {\textbf{MMS}} & {\textbf{Cohere}} & {\textbf{Qwen3}} \\
\midrule
\endfirsthead
\multicolumn{10}{c}{\tablename\ \thetable\ -- continued} \\
\toprule
& \multicolumn{3}{c}{\textbf{Ours}} & \multicolumn{6}{c}{\textbf{Baselines}} \\
\cmidrule(lr){2-4} \cmidrule(lr){5-10}
\textbf{Language}
  & {\textbf{FFT}} & {\textbf{SFT}} & {\textbf{SFTMTL}}& {\textbf{Whisper}}
  & {\textbf{Omni 1B}} & {\textbf{Omni 7B}} & {\textbf{MMS}} & {\textbf{Cohere}} & {\textbf{Qwen3}} \\
\midrule
\endhead
\midrule
\multicolumn{10}{r}{\textit{Continued on next page}} \\
\endfoot
\bottomrule
\endlastfoot
  Afrikaans & 17.39 & \underline{\textbf{16.01}} & 24.15 & 33.04 & 21.53 & 16.34 & 22.74 & 74.79 & 73.34 \\
\addlinespace[2pt]
  Amharic & \underline{\textbf{26.13}} & 91.60 & 93.33 & 154.01 & 49.93 & 30.53 & 34.10 & 128.66 & 122.97 \\
\addlinespace[2pt]
  Arabic & 32.36 & 30.39 & \textbf{27.37} & 28.56 & 30.01 & \underline{25.67} & 35.11 & 30.72 & 33.13 \\
\addlinespace[2pt]
  Armenian & \underline{\textbf{7.51}} & 40.45 & 47.72 & 55.51 & 22.53 & 15.92 & 25.91 & 145.73 & 115.86 \\
\addlinespace[2pt]
  Assamese & \textbf{35.63} & 97.03 & 97.28 & 108.02 & 36.91 & \underline{29.83} & 36.36 & 141.31 & 120.20 \\
\addlinespace[2pt]
  Asturian & 17.85 & \underline{\textbf{16.84}} & 18.56 & 51.86 & 30.79 & 23.98 & 20.11 & 63.54 & 62.96 \\
\addlinespace[2pt]
  Azerbaijani & 24.18 & \textbf{20.41} & 22.01 & 21.67 & 23.61 & \underline{17.64} & 24.58$^\dagger$ & 127.49 & 86.12 \\
\addlinespace[2pt]
  Belarusian & 62.55 & \textbf{10.67} & 11.73 & 45.75 & 15.59 & \underline{8.71} & 15.67 & 111.77 & 91.49 \\
\addlinespace[2pt]
  Bengali & \textbf{25.60} & 89.71 & 96.61 & 69.87 & 27.23 & \underline{17.41} & 33.89 & 122.12 & 116.96 \\
\addlinespace[2pt]
  Bosnian & 20.86 & \textbf{13.46} & 13.68 & 13.65 & 14.94 & \underline{9.69} & 14.43 & 108.50 & 65.75 \\
\addlinespace[2pt]
  Bulgarian & 7.12 & \underline{\textbf{4.39}} & 6.33 & 15.83 & 22.05 & 15.05 & 16.82 & 110.88 & 71.49 \\
\addlinespace[2pt]
  Burmese & \textbf{97.39} & 114.28 & 147.35 & 304.50 & 74.51 & \underline{64.65} & 100.00 & 175.14 & 137.66 \\
\addlinespace[2pt]
  Cantonese & \underline{\textbf{39.99}} & 47.53 & 45.02 & 82.80 & 102.36 & 95.24 & 100.00$^\dagger$ & 100.21 & 99.63 \\
\addlinespace[2pt]
  Catalan & 11.36 & 10.16 & 10.40 & \textbf{9.11} & 12.79 & \underline{8.83} & 12.72 & 81.96 & 71.52 \\
\addlinespace[2pt]
  Cebuano & \textbf{16.97} & 18.30 & 17.59 & 44.99 & 19.78 & 15.57 & \underline{14.47} & 105.83 & 54.85 \\
\addlinespace[2pt]
  Croatian & 18.50 & \underline{\textbf{11.67}} & 13.49 & 11.75 & 34.72 & 31.75 & 13.79 & 106.95 & 89.22 \\
\addlinespace[2pt]
  Czech & 16.01 & 12.28 & 11.68 & \textbf{10.56} & 13.25 & \underline{9.36} & 12.63 & 100.34 & 32.29 \\
\addlinespace[2pt]
  Danish & 13.23 & 17.92 & \underline{\textbf{12.60}} & 13.53 & 20.73 & 14.83 & 23.03 & 101.71 & 33.42 \\
\addlinespace[2pt]
  Dutch & 10.44 & \underline{\textbf{4.67}} & 5.82 & 5.29 & 12.49 & 10.01 & 11.98 & 11.20 & 12.82 \\
\addlinespace[2pt]
  English & 9.70 & 9.77 & 9.32 & \underline{\textbf{8.59}} & 18.15 & 13.59 & 17.21 & 21.58 & 22.05 \\
\addlinespace[2pt]
  Estonian & \underline{\textbf{5.55}} & 8.76 & 15.57 & 27.74 & 23.17 & 16.18 & 19.52 & 135.83 & 93.07 \\
\addlinespace[2pt]
  Filipino & 14.12 & 13.95 & 13.73 & \textbf{12.40} & 14.85 & \underline{12.30} & 13.00 & 105.57 & 36.70 \\
\addlinespace[2pt]
  Finnish & 26.28 & 25.03 & 24.07 & \underline{\textbf{9.29}} & 16.03 & 11.54 & 14.02 & 150.29 & 39.41 \\
\addlinespace[2pt]
  French & 17.58 & 17.54 & 17.44 & \underline{\textbf{15.99}} & 21.47 & 16.65 & 20.02 & 18.43 & 20.31 \\
\addlinespace[2pt]
  Fulah & \underline{\textbf{48.02}} & 52.00 & 53.70 & 90.27 & 53.58 & 51.82 & 49.62 & 114.22 & 97.89 \\
\addlinespace[2pt]
  Galician & 7.92 & \underline{\textbf{5.13}} & 10.73 & 14.51 & 11.76 & 7.33 & 11.23 & 61.85 & 62.69 \\
\addlinespace[2pt]
  Georgian & \textbf{39.39} & 99.97 & 129.00 & 72.57 & 19.31 & \underline{11.76} & 26.09 & 150.72 & 129.31 \\
\addlinespace[2pt]
  German & 18.63 & 9.07 & 8.21 & \underline{\textbf{7.47}} & 13.77 & 9.47 & 14.12 & 25.70 & 26.91 \\
\addlinespace[2pt]
  Greek & 30.07 & 15.91 & \underline{\textbf{10.17}} & 13.58 & 21.95 & 14.50 & 20.29 & 23.36 & 42.07 \\
\addlinespace[2pt]
  Gujarati & \textbf{31.70} & 100.45 & 99.58 & 51.68 & 23.28 & \underline{19.57} & 27.24 & 127.22 & 110.56 \\
\addlinespace[2pt]
  Hausa & 19.84 & \underline{\textbf{18.48}} & 25.59 & 89.22 & 34.13 & 27.49 & 26.00 & 112.30 & 99.77 \\
\addlinespace[2pt]
  Hebrew & 26.57 & \underline{\textbf{25.43}} & 27.79 & 28.17 & 38.98 & 29.81 & 51.86 & 120.87 & 80.21 \\
\addlinespace[2pt]
  Hindi & 36.05 & 39.57 & 38.39 & \textbf{30.98} & 59.00 & 65.71 & 27.02 & 110.86 & \underline{21.66} \\
\addlinespace[2pt]
  Hungarian & \underline{\textbf{10.21}} & 12.78 & 16.41 & 15.06 & 19.66 & 12.78 & 19.75 & 133.65 & 43.02 \\
\addlinespace[2pt]
  Icelandic & \textbf{21.38} & 32.87 & 34.86 & 38.33 & 25.58 & \underline{15.66} & 34.29 & 109.90 & 97.45 \\
\addlinespace[2pt]
  Igbo & \underline{\textbf{37.02}} & 49.56 & 48.71 & 98.91 & 60.37 & 55.21 & 46.76 & 116.76 & 98.67 \\
\addlinespace[2pt]
  Indonesian & \underline{\textbf{6.26}} & 8.34 & 6.95 & 7.02 & 18.19 & 13.36 & 15.46 & 122.29 & 19.18 \\
\addlinespace[2pt]
  Irish & \underline{\textbf{29.37}} & 31.13 & 33.28 & 119.96 & 64.13 & 52.12 & 59.98 & 110.81 & 100.37 \\
\addlinespace[2pt]
  Italian & 14.21 & 8.51 & 7.83 & \underline{\textbf{7.36}} & 11.96 & 8.20 & 10.05 & 14.64 & 16.56 \\
\addlinespace[2pt]
  Japanese & 108.72 & 72.09 & 69.89 & \underline{\textbf{51.14}} & 139.66 & 135.60 & 98.85 & 125.67 & 95.81 \\
\addlinespace[2pt]
  Javanese & \textbf{18.69} & 19.89 & 19.69 & 68.87 & 21.90 & \underline{17.84} & 20.76 & 133.31 & 75.63 \\
\addlinespace[2pt]
  Kabuverdianu & \textbf{17.67} & 17.72 & 18.75 & 89.91 & 20.36 & \underline{14.34} & 15.74 & 94.80 & 93.76 \\
\addlinespace[2pt]
  Kamba & 53.90 & \textbf{52.08} & 58.61 & 97.10 & 49.36 & \underline{40.69} & 43.48 & 134.25 & 103.32 \\
\addlinespace[2pt]
  Kannada & \textbf{39.44} & 118.47 & 119.25 & 65.69 & 23.63 & \underline{18.42} & 29.63 & 140.65 & 147.57 \\
\addlinespace[2pt]
  Kazakh & \textbf{20.09} & 37.02 & 36.52 & 39.61 & 18.02 & \underline{14.70} & 19.00 & 134.83 & 107.50 \\
\addlinespace[2pt]
  Khmer & \textbf{99.54} & 154.75 & 181.93 & 141.44 & 92.89 & \underline{86.06} & 100.00 & 686.02 & 139.78 \\
\addlinespace[2pt]
  Korean & 16.85 & \underline{\textbf{13.74}} & 15.24 & 17.23 & 35.86 & 30.69 & 54.56 & 24.33 & 21.50 \\
\addlinespace[2pt]
  Kyrgyz & 49.21 & \textbf{43.36} & 50.90 & 89.94 & 23.98 & \underline{17.01} & 20.29 & 130.02 & 107.03 \\
\addlinespace[2pt]
  Lao & \underline{\textbf{83.32}} & 99.99 & 103.32 & 111.99 & 91.07 & 84.07 & 99.96 & 602.04 & 101.14 \\
\addlinespace[2pt]
  Latvian & 16.30 & \underline{\textbf{11.60}} & 16.90 & 21.46 & 20.56 & 13.90 & 17.05 & 122.72 & 105.64 \\
\addlinespace[2pt]
  Lingala & \textbf{20.27} & 21.52 & 26.35 & 72.51 & 18.22 & \underline{14.60} & 15.69 & 116.01 & 87.93 \\
\addlinespace[2pt]
  Lithuanian & \underline{\textbf{8.35}} & 8.75 & 11.54 & 28.36 & 26.88 & 17.88 & 18.41 & 124.71 & 112.68 \\
\addlinespace[2pt]
  Luganda & 65.92 & \textbf{44.80} & 59.41 & 116.82 & 39.98 & \underline{30.89} & 33.55 & 128.74 & 114.99 \\
\addlinespace[2pt]
  Luo & 82.06 & \textbf{77.47} & 79.12 & 92.67 & 38.69 & \underline{31.98} & 38.32 & 107.41 & 99.00 \\
\addlinespace[2pt]
  Luxembourgish & \textbf{25.41} & 35.65 & 36.39 & 85.46 & 38.26 & \underline{25.36} & 32.43 & 96.69 & 95.68 \\
\addlinespace[2pt]
  Macedonian & 7.95 & \underline{\textbf{4.94}} & 7.16 & 20.77 & 13.65 & 9.73 & 11.86 & 109.31 & 41.34 \\
\addlinespace[2pt]
  Malay & 12.16 & 9.49 & 9.06 & \underline{\textbf{7.62}} & 18.22 & 13.23 & 16.33$^\dagger$ & 136.11 & 28.46 \\
\addlinespace[2pt]
  Malayalam & \textbf{48.48} & 99.41 & 99.46 & 119.86 & 45.42 & 37.87 & \underline{37.71} & 139.06 & 149.48 \\
\addlinespace[2pt]
  Maltese & 11.85 & \underline{\textbf{11.28}} & 14.28 & 79.46 & 39.62 & 30.52 & 17.88 & 110.68 & 105.43 \\
\addlinespace[2pt]
  Mandarin & \underline{\textbf{31.51}} & 72.75 & 74.93 & 53.63 & 97.27 & 88.54 & 100.00$^\dagger$ & 97.69 & 99.98$^\dagger$ \\
\addlinespace[2pt]
  Maori & \underline{\textbf{19.73}} & 24.65 & 25.50 & 40.05 & 28.95 & 22.42 & 25.09 & 107.62 & 83.63 \\
\addlinespace[2pt]
  Marathi & 64.56 & \textbf{59.29} & 72.07 & 61.96 & 33.41 & \underline{26.23} & 34.27 & 124.76 & 89.29 \\
\addlinespace[2pt]
  Mongolian & \underline{\textbf{13.34}} & 26.46 & 29.88 & 88.20 & 46.24 & 30.50 & 35.68 & 117.86 & 105.39 \\
\addlinespace[2pt]
  Nepali & \textbf{36.25} & 45.85 & 52.46 & 52.78 & 37.78 & 32.75 & \underline{32.35}$^\dagger$ & 128.47 & 110.49 \\
\addlinespace[2pt]
  Northern Sotho & \textbf{29.40} & 37.94 & 40.14 & 111.17 & 35.63 & \underline{22.17} & 27.24 & 116.30 & 99.73 \\
\addlinespace[2pt]
  Norwegian & 9.64 & 9.20 & \underline{\textbf{6.86}} & 14.77 & 20.43 & 17.78 & 24.23 & 102.33 & 68.10 \\
\addlinespace[2pt]
  Nyanja & \textbf{40.68} & 47.24 & 47.05 & 112.10 & 41.50 & \underline{31.97} & 34.22 & 155.85 & 115.10 \\
\addlinespace[2pt]
  Occitan & \underline{\textbf{26.87}} & 34.41 & 30.61 & 70.53 & 46.66 & 38.10 & 37.07 & 88.24 & 85.76 \\
\addlinespace[2pt]
  Oriya & \textbf{39.40} & 112.22 & 107.14 & 107.73 & 45.74 & 40.36 & \underline{37.86} & 130.08 & 125.94 \\
\addlinespace[2pt]
  Oromo & \underline{\textbf{59.83}} & 71.60 & 72.27 & 100.84 & 74.62 & 70.76 & 64.71 & 104.87 & 109.41 \\
\addlinespace[2pt]
  Pashto & 37.64 & \textbf{31.40} & 41.18 & 89.67 & 32.61 & \underline{28.76} & 51.84 & 105.61 & 96.42 \\
\addlinespace[2pt]
  Persian & \underline{\textbf{14.45}} & 33.67 & 34.48 & 37.66 & 28.45 & 22.02 & 26.06 & 114.00 & 44.16 \\
\addlinespace[2pt]
  Polish & 23.31 & 11.34 & 10.73 & \underline{\textbf{7.10}} & 13.68 & 10.35 & 11.88 & 25.11 & 30.17 \\
\addlinespace[2pt]
  Portuguese & 10.78 & 9.76 & \underline{\textbf{8.17}} & 8.46 & 15.88 & 12.59 & 13.50 & 20.03 & 22.24 \\
\addlinespace[2pt]
  Punjabi & \underline{\textbf{24.50}} & 78.30 & 80.76 & 65.90 & 39.13 & 40.68 & 31.21 & 121.58 & 101.06 \\
\addlinespace[2pt]
  Romanian & 15.35 & 11.35 & \underline{\textbf{10.21}} & 11.08 & 15.84 & 12.48 & 12.10 & 105.46 & 36.24 \\
\addlinespace[2pt]
  Russian & 16.04 & 9.24 & 8.74 & \underline{\textbf{7.91}} & 19.01 & 12.34 & 24.00 & 115.37 & 22.73 \\
\addlinespace[2pt]
  Serbian & \underline{\textbf{44.46}} & 74.23 & 50.08 & 71.85 & 48.49 & 45.06 & 99.29$^\dagger$ & 108.02 & 89.40 \\
\addlinespace[2pt]
  Shona & \textbf{34.31} & 38.66 & 41.09 & 116.55 & 23.37 & \underline{18.19} & 22.97 & 164.22 & 124.23 \\
\addlinespace[2pt]
  Sindhi & \textbf{25.48} & 39.10 & 39.51 & 182.90 & 28.46 & \underline{21.33} & 24.90 & 125.92 & 101.32 \\
\addlinespace[2pt]
  Slovak & \underline{\textbf{6.77}} & 17.63 & 10.54 & 17.73 & 22.54 & 18.61 & 11.92 & 100.32 & 74.20 \\
\addlinespace[2pt]
  Slovenian & 12.29 & 7.83 & \underline{\textbf{6.93}} & 20.75 & 25.54 & 21.48 & 16.71 & 106.80 & 95.72 \\
\addlinespace[2pt]
  Somali & \textbf{50.96} & 55.80 & 56.86 & 91.82 & \underline{38.88} & 39.33 & 44.47 & 109.51 & 103.67 \\
\addlinespace[2pt]
  Sorani Kurdish & \underline{\textbf{26.43}} & 48.14 & 62.31 & 108.08 & 39.51 & 29.99 & 39.09 & 110.71 & 105.26 \\
\addlinespace[2pt]
  Spanish & 10.82 & 5.25 & 5.41 & \underline{\textbf{4.61}} & 8.76 & 6.23 & 8.66 & 14.20 & 15.08 \\
\addlinespace[2pt]
  Swahili & \textbf{18.33} & 22.38 & 27.24 & 58.04 & 20.93 & \underline{17.03} & 20.69 & 107.38 & 75.04 \\
\addlinespace[2pt]
  Swedish & \underline{\textbf{6.55}} & 9.67 & 6.73 & 8.16 & 20.55 & 14.42 & 21.16 & 108.08 & 29.20 \\
\addlinespace[2pt]
  Tajik & 19.96 & \textbf{18.96} & 22.10 & 87.57 & 15.33 & \underline{13.94} & 16.39 & 112.10 & 107.03 \\
\addlinespace[2pt]
  Tamil & 81.68 & 77.13 & 98.05 & \textbf{46.54} & 43.76 & \underline{35.86} & 43.80 & 126.50 & 140.03 \\
\addlinespace[2pt]
  Telugu & \textbf{45.49} & 121.65 & 105.66 & 101.25 & 36.80 & \underline{32.29} & 40.02 & 131.08 & 133.47 \\
\addlinespace[2pt]
  Thai & 114.72 & 100.92 & 107.00 & \underline{\textbf{50.00}} & 77.84 & 67.45 & 94.12 & 711.64 & 69.32 \\
\addlinespace[2pt]
  Turkish & 14.55 & 12.44 & 13.07 & \underline{\textbf{11.68}} & 25.71 & 19.14 & 22.97 & 127.12 & 25.87 \\
\addlinespace[2pt]
  Ukrainian & 31.08 & 13.22 & \underline{\textbf{10.60}} & 11.60 & 21.84 & 14.56 & 22.07 & 109.58 & 84.15 \\
\addlinespace[2pt]
  Umbundu & 56.28 & \textbf{54.36} & 59.53 & 134.62 & 52.89 & 51.70 & \underline{43.66} & 143.41 & 130.79 \\
\addlinespace[2pt]
  Urdu & 27.96 & 26.77 & 27.03 & \underline{\textbf{22.81}} & 78.03 & 99.67 & 30.97 & 112.60 & 100.44 \\
\addlinespace[2pt]
  Uzbek & 22.09 & \underline{\textbf{16.00}} & 27.19 & 89.91 & 35.63 & 28.61 & 27.89$^\dagger$ & 122.65 & 116.34 \\
\addlinespace[2pt]
  Vietnamese & 12.96 & 10.68 & 10.95 & \underline{\textbf{7.51}} & 17.44 & 12.57 & 30.33 & 20.26 & 17.52 \\
\addlinespace[2pt]
  Welsh & \underline{\textbf{12.57}} & 15.15 & 15.89 & 34.59 & 41.88 & 25.73 & 35.63 & 112.01 & 102.72 \\
\addlinespace[2pt]
  Wolof & \textbf{43.89} & 46.25 & 47.45 & 132.37 & 41.15 & 39.20 & \underline{39.13} & 120.71 & 98.88 \\
\addlinespace[2pt]
  Xhosa & \textbf{47.87} & 51.72 & 53.26 & 136.67 & 44.30 & \underline{35.41} & 36.71 & 206.41 & 139.66 \\
\addlinespace[2pt]
  Yoruba & \textbf{59.98} & 70.26 & 76.02 & 101.27 & 52.34 & \underline{50.25} & 53.52 & 121.71 & 101.21 \\
\addlinespace[2pt]
  Zulu & \textbf{41.40} & 48.45 & 49.50 & 136.14 & 35.46 & \underline{29.50} & 33.57 & 188.94 & 138.03 \\
\end{longtable}
\twocolumn

\end{document}